\documentclass[11pt,a4paper]{article}
\usepackage{times,latexsym}
\usepackage{url}
\usepackage[T1]{fontenc}
\usepackage{hyperref}
\usepackage{graphicx,dblfloatfix}
\usepackage{colortbl}
\usepackage{amsmath}
\usepackage{comment}
\usepackage{subcaption}
\usepackage{float}

\usepackage[acceptedWithA]{tacl2021v1}

\usepackage{xspace,mfirstuc,tabulary}

\newif\iftaclinstructions
\taclinstructionsfalse %
\iftaclinstructions
\renewcommand{\confidential}{}
\renewcommand{\anonsubtext}{(No author info supplied here, for consistency with
TACL-submission anonymization requirements)}
\newcommand{\instr}
\fi

\iftaclpubformat %

\else

\fi

\title{As Easy as Rocket Science: Assessing the Ability of Large Language Models to Interpret Negation in Figurative Language  }

\author{
 Jasmine Owers\thanks{~~Corresponding authors: JO (jo16726@bristol.ac.uk) or  ML (m.a.f.lewis@uva.nl)}\\
 Intelligent Systems Lab
 \\
 University of Bristol\\UK
 \\
 \And 
 Edwin Simpson
 \\
 Intelligent Systems Lab
 \\
 University of Bristol\\UK
 \\
   \And
 Martha Lewis$^*$\\
 ILLC
 \\
 University of Amsterdam
 \\
 The Netherlands
}

\date{}

\begin{document}
\maketitle
\begin{abstract}
Figurative language and negation are two areas that challenge current language models, however, both are widely used throughout written and spoken language. Large language models (LLMs) are also widely used in everyday contexts where they cannot necessarily be tuned for a specific dataset. It is therefore essential to understand the ability of LLMs to correctly interpret text that includes both negation and figurative language. To investigate this, we develop a set of new annotations to an existing dataset of figurative language, and test a range of language models on the dataset. We find that the combination of negation and figurativeness can present a particular challenge, and that performance overall and across different negation types is particularly dependent on the prompt style used.

\end{abstract}

\section{Introduction}

Figurative language is prevalent throughout everyday language use \cite{veale_metaphor_2016,roberts_why_1994}. Metaphor, in particular, has been argued to form a fundamental foundation to our cognitive and creative abilities \cite{lakoff_80}. As such, language models need to be able to appropriately respond to uses of figurative language in multiple downstream contexts such as text summarization, natural language understanding (NLU), and natural language inference (NLI). Furthermore, models should be able to do this out-of-the-box, without involving specific finetuning. 

Previous research has shown that figurative language can cause difficulty for language models, particularly in the area of NLI \citep{agerri_08,chakrabarty_21}. On the other hand, positive performance for language models in this domain is seen in \citet{stowe_22,liu_22}. In these cases, the models are tested on mostly conventional metaphor \citep{stowe_22}, or good performance is attained via fine-tuning \citep{liu_22}. Much work on figurative language and NLI has concentrated on metaphor; one exception is \citet{stowe_22} who also examine the effect of sarcasm, although only in contradictory contexts.

Negation is also a known challenge for models to interpret, especially without specific finetuning \citep{hossain_20}, and research into performance on text combining both negation and figurative language is limited. There is thus a gap in understanding the ability of modern language models to correctly interpret novel figurative language in contexts involving negation. To address this gap, we provide a set of novel annotations for the Fig-QA dataset \cite{liu_22}, and assess the ability of a range of language models to interpret novel figurative language in combination with negation. We analyse the embedding space of one model to discover other linguistic features that may contribute to performance in metaphor interpretation, and find that tense and concreteness also play a role, although somewhat outweighed by the effect of negation. To understand the effect of negation in the absence of figurative language, we develop a small literal paraphrasing dataset, finding that models tend to perform better in the literal setting. We therefore find that the combination of negation and figurative language is particularly challenging for a range of language models. This paper makes the following contributions:

\begin{itemize}
\item We provide annotations for metaphor/simile, negation, tense and concreteness to Fig-QA \citep{liu_22}, an existing dataset of less conventional figurative language.
\item We create a new small-scale dataset of literal negation based on the Fig-QA dataset, designed to isolate the effects of figurative language and negation.
\item We test a wide range of models, including recent LLMs, on these datasets and give an analysis of the embedding space of one model on this task.
\item We find that the combination of negation and figurativeness presents a particular challenge, and that models' performance is sensitive to the type of prompt used.
\end{itemize}
Code and data are available at %
\url{https://github.com/jrdowers/Negation-and-Fig-Lang}.

\section{Background and Related Work}
\subsection{Types of Figurative Language}
In this paper, we consider metaphor and simile, as well as irony, in the form of simile combined with negation. Both metaphor and simile provide an explanation of one concept in terms of another. In the case of metaphor, the concept to be explained (the target) is more complex or abstract, and is explained in terms of a simpler or more concrete concept (the source). For example, in \textit{the fire \underline{devoured} the home}, the properties of the source domain (literal meaning of devour) are applied to the target domain (destruction) to enhance the meaning \citep{lakoff_80}. In the case of simile, an explicit comparison--often with exaggeration--is given using words such as “like” or “as”. In \textit{the box was \underline{as light as a feather}} the box is described in terms of the lightweight property of a feather.

Both metaphor and simile are prevalent in natural language (for metaphor, see statistics calculated in \citet{shutova_10, steen_10}), and metaphor is argued to be driven by an underlying cognitive process by which abstract or unfamiliar concepts are understood by imagining them as more concrete concepts \cite{lakoff_80}. Research in the field of human psychology and cognitive science shows that, overall, people are able to understand metaphor as easily as literal sentences \citep{cacciari_94}, and we therefore argue that if language models are to behave in a human-like way, they should also be able to understand metaphor easily.

\subsection{Figurative Language Interpretation}

One approach to assessing figurative language interpretation is by assessing the ability of a model to correctly paraphrase a sentence that uses metaphor \citep{shutova_10}. This approach is taken in \citet{bizzoni_predicting_2018,tong_24,liu_22}, where metaphorical sentences are paired with two or more paraphrases, one of which is apt, and the task of the model is to pick out the apt paraphrase. In general, models do not perform at human level at this task, although \citet{liu_22} find that with fine-tuning the performance of RoBERTa improves considerably.

Figurative language interpretation has also been assessed via a natural language inference task 
\cite{chakrabarty_21, stowe_22}. Related tasks include asking models to generate explanations \citep{chakrabarty_22_flute} or classifying text continuations \citep{chakrabarty_22}. Fine-tuning again helped performance, as did enhancing models with a knowledge-base component in \citet{chakrabarty_22}.

Recent work has shown that newer models are improving at metaphor interpretation and in particular novel metaphor. \citet{ichien_24} show that GPT-4 achieves close to human performance on Fig-QA and is also able to generate impressive interpretations of novel metaphors. The prompt used for testing GPT-4 gave the figurative sentence and a choice of two literal sentences. In this paper, we probe this further by analysing performance at a more fine-grained level.

\begin{table*}[t]
  \centering
    \footnotesize
    \begin{tabular}{rlll}
          & & Figurative & Interpretation \\
    \hline
    \multicolumn{1}{l}{metaphor pair} & & This job is a field of flowers. & The job is easy and nice. \\
          & & This job is a hammer to the knees. & The job is brutal and difficult. \\
    \hline
    \multicolumn{1}{l}{\textit{non-neg} pair} & & The classroom was like a freezer. & The classroom was really cold. \\
          & & The classroom was like an oven. & The classroom was really hot. \\
    \hline
    \multicolumn{1}{l}{\textit{not} pair} & intens & This is as fun as a bowl of nachos. & It's very fun. \\
          & neg & This is as fun as stubbing your toe. & It's not fun at all. \\
    \hline
    \multicolumn{1}{l}{\textit{antonym} pair} & intens & She's as gentle as a puppy. & She's very gentle. \\
          & neg & She's as gentle as a shark. & She's vicious. \\
    \end{tabular}%
  \caption{Examples of different types of metaphor and simile instances in Fig-QA}
  \label{tab:simile example}
\end{table*}
\subsection{Negation}
In natural language, between 9\% and 30\% of sentences contain negation depending on domain \citep{hossain_20}. 
As negation is so prevalent, it is crucial for language models to be able to accurately interpret it.
However, negation is also known to present a particular challenge to models in NLU. \citet{hossain_20,kassner_20,garcia-ferrero_23}, show that models have trouble interpreting negation, with some improvement after finetuning \cite{hossain_20,hosseini_21}. Models can rely on surface patterns when reasoning about sentences including negation; \citet{garcia-ferrero_23} show that in a True/False classification task, models are more likely to label sentences including negation as False. Further, the pretraining of the model can affect this; \citet{truong_23} found that instruction-tuned models perform better with negation, particularly where they are prompted to reason about negation.

\subsection{Combining Figurative Language and Negation}
All of the above points to the idea that language models may have difficulty in combining figurative language and negation. Indeed, this has been briefly noted in \citet{liu_22}, and negated sentences are included in \citet{stowe_22}.
In this paper, we further investigate the interaction of figurative language and negation. We examine the effect of negation on the accuracy of metaphor interpretation in a paraphrasing task, look at the impact of different prompting styles, and compare a range of models. Experiments do not include fine-tuning -- we argue that in particular large pretrained LLMs should be able to handle the extremely prevalent phenomena of negation and figurative language out-of-the-box.

\section{Negation Dataset} \label{dataset}

To understand the effect of metaphor, simile, and negation in NLU, we develop a set of annotations for the Fig-QA dataset, originally created to support interpretation of figurative language. This dataset consists of 10,256 instances of figurative sentences with literal interpretations, generated by crowdworkers who were instructed to generate pairs of rare or creative, but easily interpretable metaphors with opposite meanings, along with literal paraphrases. Each figurative sentence has one correct and one incorrect paraphrase. Most figurative sentences in the dataset are similes, and most metaphorical sentences are copula metaphors\footnote{metaphors of the form \textit{X} is a \textit{Y}, e.g.\ `The girl is a mouse'.}. 
The full dataset of 10,256 sentences was split by the original authors into a training set of 9,110 sentences and a test set of 1,146 sentences.

We create an approximate classification of the instances into metaphors (\textit{met}) (e.g.\ ``When he hid, he became a mouse'') and similes (\textit{sim}) by classifying similes as containing ``as'' (e.g.\ ``The display was bright as the sun''), ``like'' (e.g.\ ``The joke is like a ton of concrete'') or ``the * of'', (e.g.\ ``The runner had the speed of a fired bullet '').

Within the simile category, the paraphrases generally take the form ``NOUN is very ADJ'', where the simile is an intensification of the adjective, or ``NOUN is not ADJ at all'', where the simile negates the adjective. We therefore categorise the simile instances into separate cases where the simile acts either as an intensification (\textit{intens}) or a negation (\textit{neg}). Similes generally come in pairs, where exactly one is classed as \textit{neg} and the other as \textit{intens}. We call these pairs \textit{neg pairs}, and the remaining non-paired similes we call \textit{non-neg pairs}.

Within the category of negated similes, we define two types of negating similes: \textit{not}, containing either ``not'', ``n't'' or ``no'' in exactly one of the two interpretations, and \textit{antonym}, containing the structure ``as * as'' in the simile, and the adjective, ``*'', appearing in exactly one of the two interpretations. If the ``as * as'' structure contains a multi-word phrase, the first word is used, e.g.\ ``easy'' in ``as easy to read as'', see examples in Table \ref{tab:simile example}.

Table \ref{tab:figqa counts} shows the number of instances in each category. Labels are provided for the 9,110 sentences included in the Fig-QA training set; each of these sentences has been paired with two possible paraphrases, and a label is available for which paraphrase is correct. We use this larger set of annotated sentences in most of our analyses. The total of \textit{neg pairs} and \textit{non-neg pairs} is 7064; there are 106 simile instances not included in these: 90 similes whose partner is a metaphor, 7 simile pairs (14 instances) with inconsistent formats and two unpaired sentences. There are 4 more \textit{intens} instances than \textit{neg} instances; these are two pairs in which the adjective is different in each sentence: ``The cake was as \textit{hard} as a rock.'' but ``The cake was as \textit{light} as a feather'', and ``The book had a plot that many would find as \textit{Boring} as watching paint dry'' but ``The book had a plot that many would find as \textit{Exciting} as a three-ring circus''. Thus, both sentences in the pair are intensifications. These are also the two extra pairs in \textit{intens ``antonym''}.

We manually validated the annotations of 184 sentences from the train set (2\% of the dataset) randomly selected proportionally to each category. We found that 7 of the sentences in the sample were incorrectly annotated: 3 metaphors were classified as similes (non-neg pairs), 2 similes (non-neg pairs) were classified as metaphors, 1 simile was incorrectly classified as a neg pair and 1 sentence was broken in the original dataset. Since our key analyses involve similes that are in the \textit{neg pairs} subset of the dataset, these misclassifications have minimal impact.

\begin{table}
    \centering
      \footnotesize
\begin{tabular}{l|r}
\multicolumn{1}{l}{Data split} & \multicolumn{1}{l}{Count} \\
\hline
all   & 9110 \\
\hspace{3mm}met   & 1940 \\
\hspace{3mm}sim   & 7170 \\
\hspace{6mm}non-neg pairs & 3762 \\
\hspace{6mm}neg pairs & 3302 \\
\hspace{9mm}neg   & 1649 \\
\hspace{12mm}neg ``not'' & 1029 \\
\hspace{12mm}neg ``antonym'' & 620 \\
\hspace{9mm}intens & 1653 \\
\hspace{12mm}intens ``not'' & 1029 \\
\hspace{12mm}intens ``antonym'' & 624 \\
\end{tabular}%
\caption{Number of sentences in each split of the Fig-QA train set}
\label{tab:figqa counts}
\end{table}

\section{Methods}

The task of Fig-QA is as follows: given a triple of sentences (\textit{fig}, \textit{lit$_1$}, \textit{lit$_2$}), a model must select the most appropriate paraphrase out of \textit{lit$_1$} and \textit{lit$_2$}. One of \textit{lit$_1$} and \textit{lit$_2$} is a literal paraphrase of \textit{fig}, and the other contradicts it (recall Table \ref{tab:simile example} for examples). We test a range of different models including embedding-based models and open- and closed-source LLMs. The range of models necessitates a range of different assessment methods. For embedding models, we compare the cosine similarity of the embeddings of the figurative sentence and each paraphrase. However, autoregressive models such as Llama do not have a natural sentence embedding output as they are trained to output probabilities over sequences. We therefore use the following two methods for autoregressive models: we use LLM logprobs, where available, to determine the likelihood of each paraphrase to follow the figurative sentence; and we prompt models with the figurative sentence and both paraphrases and ask which is most appropriate. Where possible we test each autoregressive model on both methods.

\subsection{Models}
We test a series of models ranging from simple to state of the art:
GloVe \citep{glove},
SBERT (all-mpnet-base-v2 accessed via Hugging Face) \citep{sbert},
Llama-3, -3.1 and -3.3 \citep{llama3},
GPT-4o-mini,
GPT-4o \citep{gpt-4o} and OpenAI o1-mini \citep{openai-o1}.\footnote{We also ran tests on Word2Vec \citep{mikolov_2013}, ms-Word2DM-d10 \citep{meyer_20}, Infersent \citep{infersent}, and SBERT all-MiniLM-L6-v2; results for these models are available at \url{https://github.com/jrdowers/Negation-and-Fig-Lang}}
Llama models were accessed via the Together AI API\footnote{https://api.together.ai/} and GPT models were accessed via the OpenAI API\footnote{https://platform.openai.com/}.

All Llama 3 models were quantised to FP8 as these were the versions of the models provided by Together AI. GloVe was tested on a lemmatised version of the sentences since the GloVe embeddings we used did not have every unlemmatised word in the vocabulary. The auto-regressive LLMs were tested using the full unlemmatised sentences from the dataset.

We also tested Llama-2 \citep{llama2}, quantised to FP4, and GPT-Neo \citep{gpt-neo}, accessed via Hugging Face and run on a local server. These results can be found in Appendix \ref{appx:original_results}.

\begin{figure*}
    \centering
    \begin{subfigure}{0.4\textwidth}
    \includegraphics[height=0.25\textheight]{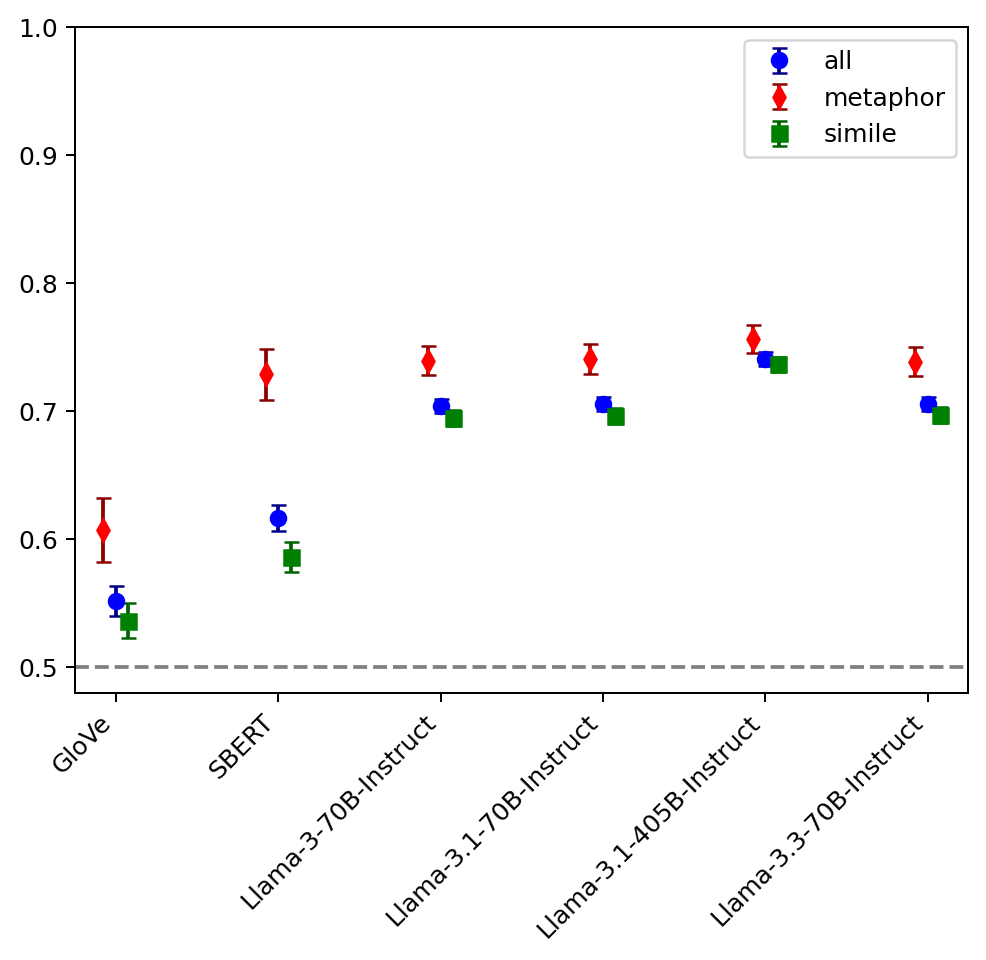}
    \caption{Embedding and Mid-Phrase}
    \label{fig:figqa_results_overall_em}
    \end{subfigure}
    \hspace{1cm}
    \begin{subfigure}{0.4\textwidth}
    \includegraphics[height=0.25\textheight]{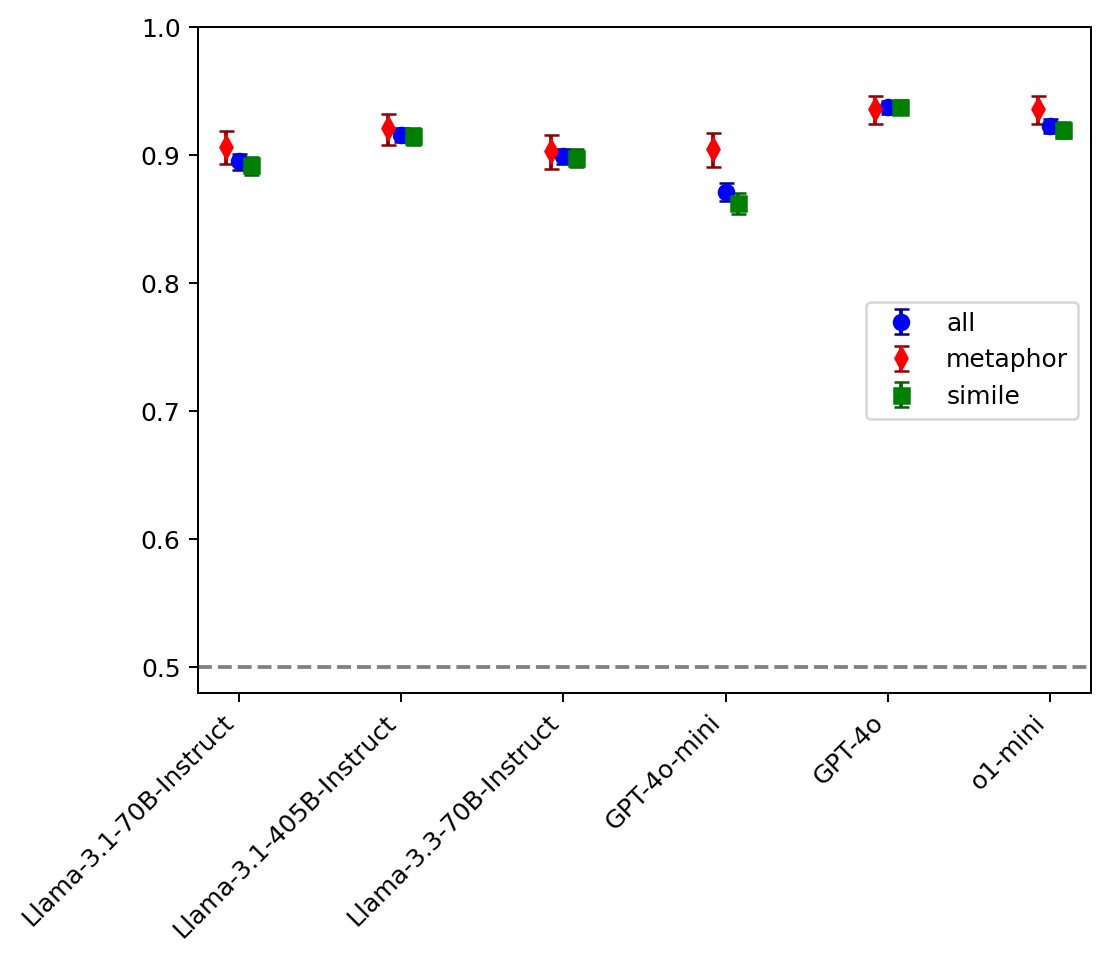}
    \caption{Question-Answer}
    \label{fig:figqa_results_overall_qa}
    \end{subfigure}
    \caption{Accuracy of models on Fig-QA train set using (a) the embedding and mid-phrase methods and (b) the question-answer method. The bars indicate 95\% binomial confidence intervals except Llama models in (a) which are 95\% bootstrap confidence intervals. The dotted line is the random baseline.}
    \label{fig:figqa results overall}
\end{figure*}
\subsection{Scoring Methods} \label{method-comp}
\textbf{Embedding}. To test the performance of GloVe and SBERT, we use cosine similarity of embeddings. A sentence embedding is generated for each sentence in the triple (\textit{fig}, \textit{lit$_1$}, \textit{lit$_2$}). In the case of GloVe, this is achieved by element-wise addition of the word embeddings. The cosine similarity is calculated between the vector for \textit{fig} and the vectors for each of \textit{lit$_1$} and  \textit{lit$_2$}. The model then selects the paraphrase with the highest similarity. Each figurative sentence is classified as paraphrased correctly or incorrectly.

Two methods are used for the auto-regressive models: a \textbf{Mid-Phrase} method and a \textbf{Question-Answer} method. The \textbf{Mid-Phrase} method compares a normalised log-likelihood of two candidate phrases. First, the input phrases are prepared. These are the figurative sentence, followed by a mid-phrase such as ``In other words,'', followed by one of the candidate paraphrases, for example ``\textit{She’s as gentle as a puppy. In other words, She's very gentle.}''
. When adding the mid-phrase, we standardise the punctuation at the end of the figurative sentence and the capitalisation of the paraphrase. We then calculate a length-normalised phrase log-likelihood $L$. This is calculated as the mean of the token log-likelihoods in the phrase, or equivalently the negative log of the per-token perplexity as defined in \citet{jurafsky_25}.
\begin{equation}
    L = \frac{1}{N} \sum_{i=1}^{N} \ln P(w_i \mid w_1, w_2, \ldots, w_{i-1})
    \label{eq:phrase_likelihood}
\end{equation}
Each figurative sentence is classified as correctly paraphrased if %
$L$ for the apt paraphrase is higher than for the inapt, and incorrect otherwise.

We test 5 mid-phrases: ``That is to say, '', ``In other words, '', ``Put differently, '', a space and a newline.

In the \textbf{Question-Answer} method, the model is prompted with the figurative sentence and both paraphrases, and asked which is the best paraphrase. The prompt format is as follows:

`Which of the following sentences, sentence A or sentence B, most accurately describes the meaning of the following: ``Her word had the strength of titanium.''? Sentence A: ``Her promises can be believed.'' Sentence B: ``Her promises cannot be trusted.''\symbol{92}n\symbol{92}nAnswer: Sentence '

For GPT models, the next generated token is used (`A' or `B'). To account for model bias towards choosing the first answer presented, for every other pair of figurative sentences the order of the paraphrases in the prompt was reversed.

For Llama models the output was inconsistent, generating varied chatty responses. 
Furthermore, the model almost always picked the first paraphrase, sentence A. We therefore use a log-likelihood based method for the Llama models. We use equation \eqref{eq:phrase_likelihood} to calculate the log-likelihood of the phrase with each of the answers `A' or `B', appended to the end of the prompt and select the answer with the highest log-likelihood. This method is equivalent to determining which output is more likely, `A' or `B'.

We include human performance on the test split of the dataset, provided by \citet{liu_22}. Annotators saw both paraphrases and chose the most appropriate; this is equivalent to our question-answer method. Since human performance is only available on the test set, we also tested some models on the test set for a direct comparison (results are shown in Appendix \ref{appx:test_results}).
There is a 0.2\% difference between our calculated overall accuracy of human annotators and that reported in \citet{liu_22}.

For all methods, the task is a binary classification, we measure performance using accuracy, and the random baseline is 0.5. To test statistical significance we use McNemar’s test where the data can be paired and a one-sided two-proportion $z$-test otherwise, with Benjamini-Hochberg correction for multiple comparisons.

\section{Negation Results}
We test models on each of the 5 mid-phrases. The mean of the accuracy using the connector mid-phrases (``That is to say, '', ``In other words, '', ``Put differently, ''), and the connector-free mid-phrases (a space and a newline) for the Fig-QA train set are shown in Table \ref{tab:c_vs_cf}. The connector-free mid-phrases consistently result in significantly lower accuracy ($p<<0.01$). This is also consistent in each split of the dataset considered below, so we only report the average of connector mid-phrases for the rest of the paper. Results for each individual mid-phrase can be found in Appendix \ref{appx:midphrase}.

As a first analysis, we calculate the performance of models overall and split by metaphor and simile (Figure \ref{fig:figqa results overall}).  Across every model,  on the train set, performance on metaphorical sentences was greater than or equal to performance on sentences involving similes. Human performance on the test set was equal across sentence types (all: 0.946, 95\% binomial confidence interval [0.931-0.958], met: 0.943 [0.907-0.966], sim: 0.946 [0.930-0.960]). Models using the question-answer method were close to human performance on the test set, Llama3.1-405B and GPT-4o showing no significant difference from human performance (Figure \ref{fig:figqa_results_overall_qtest}, Appendix \ref{appx:test_results}). Furthermore, performance for a given model using the question-answer prompt style (Figure \ref{fig:figqa_results_overall_qa}) was stronger than performance when using the mid-phrase method (Figure \ref{fig:figqa_results_overall_em}).

\begin{table*}[h]
  \centering
    \footnotesize
      \begin{tabular}{l|llll}
      & Llama-3-70B-Instruct & Llama-3.1-70B-Instruct & Llama-3.1-405B-Instruct & Llama-3.3-70B-Instruct \\
      \hline
      connector & 0.7043 [0.6989-0.7097] & 0.7060 [0.7005-0.7114] & 0.7407 [0.7355-0.7459] & 0.7058 [0.7004-0.7112] \\
      connector-free & 0.6529 [0.6459-0.6597] & 0.6513 [0.6443-0.6582] & 0.6685 [0.6616-0.6753] & 0.6461 [0.6392-0.6531] \\
      \end{tabular}%
    \caption{Overall accuracy on Fig-QA using the mid-phrase method showing average across connector and connector-free mid-phrases, with 95\% bootstrap confidence intervals}
    \label{tab:c_vs_cf}
\end{table*}
\label{figqa analysis 2}
\begin{figure*}
    \centering
    \begin{subfigure}{0.4\textwidth}
    \includegraphics[height=0.25\textheight]{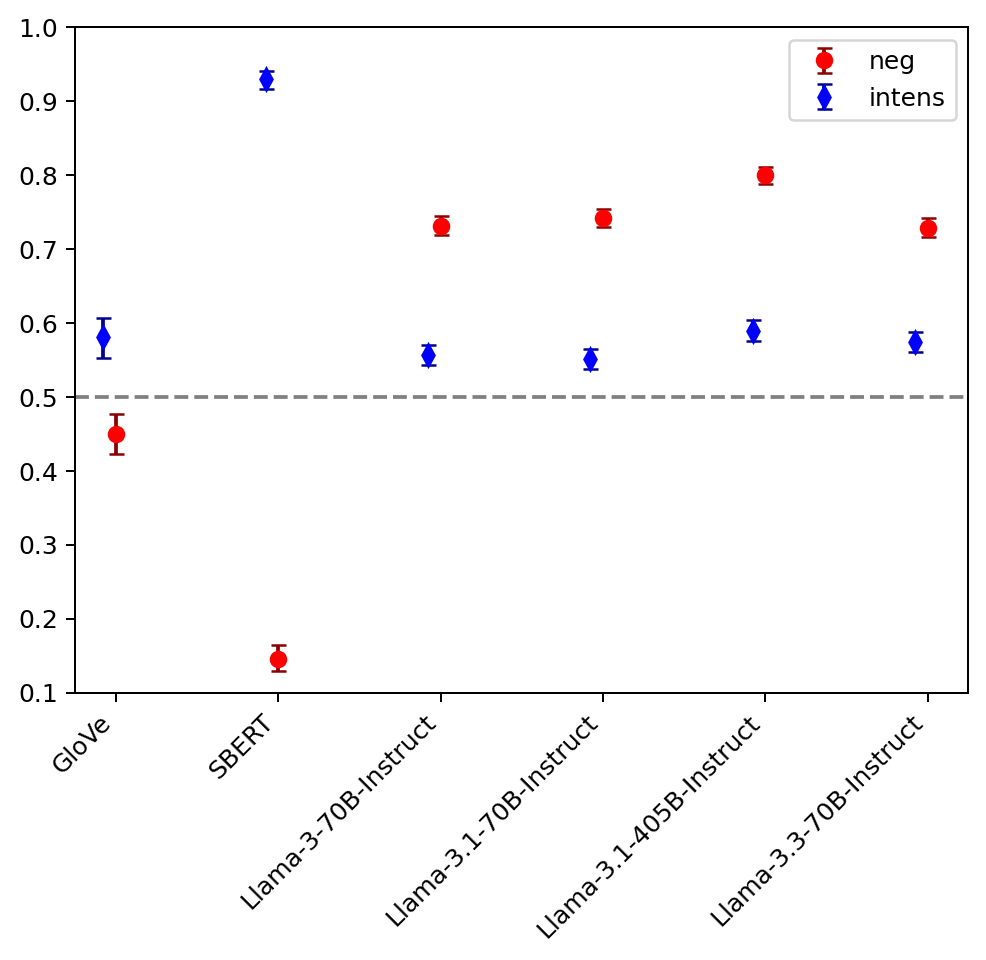}
    \caption{Embedding and Mid-Phrase}
    \label{fig:figqa_results_neg_em}
    \end{subfigure}
    \hspace{1cm}
    \begin{subfigure}{0.4\textwidth}
        \includegraphics[height=0.25\textheight]{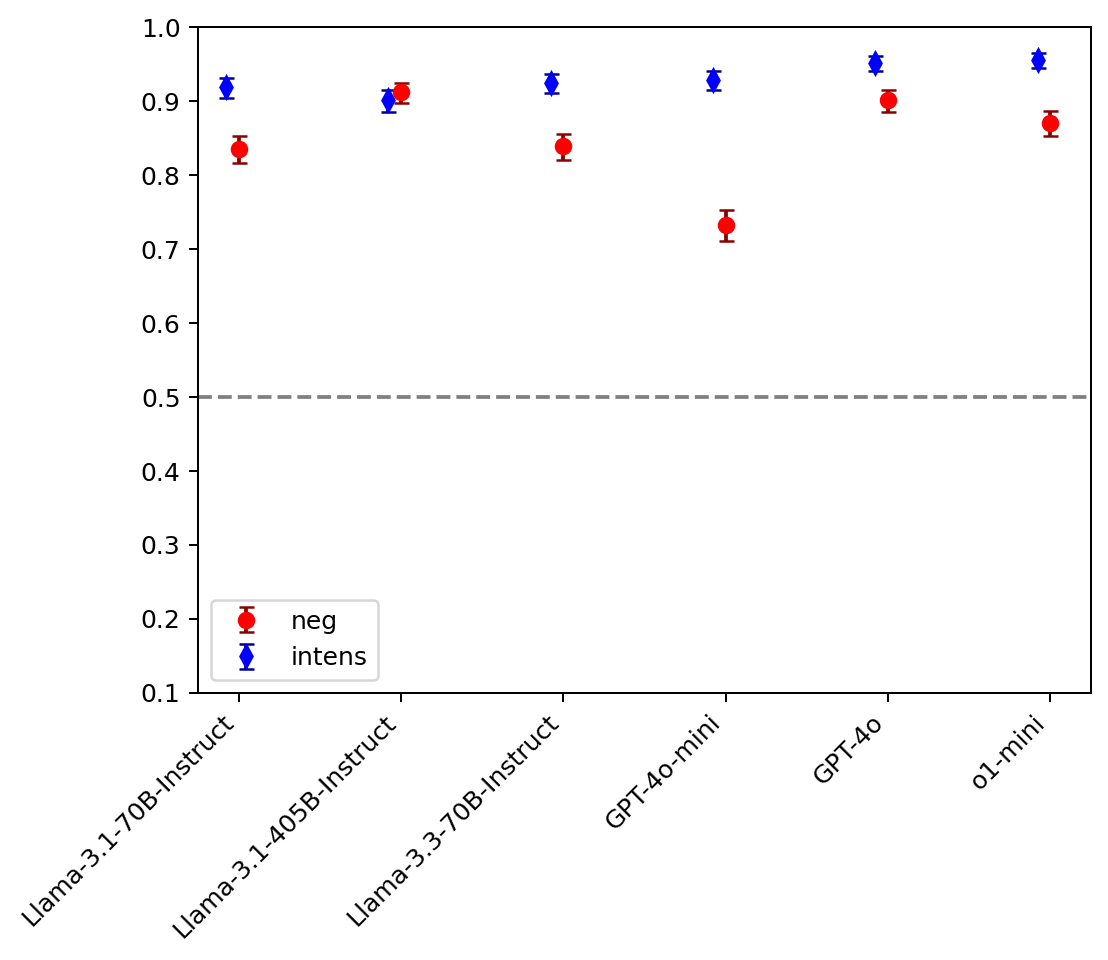}
        \caption{Question-Answer}
        \label{fig:figqa_results_neg_qa}
    \end{subfigure}
    \caption{Accuracy of models on negating and intensifying similes in the Fig-QA train set using (a) the embedding and mid-phrase methods and (b) the question-answer method. The bars indicate 95\% binomial confidence intervals except Llama models in (a) which are 95\% bootstrap confidence intervals. The dotted line is the random baseline. Note that some models achieve a higher accuracy with negating similes than intensifying.}
    \label{fig:figqa results neg}
\end{figure*}

\subsection{Negating vs Intensifying Similes}

We conjecture that models perform more weakly on simile due to the presence of negation, as negation is more common in similes than metaphors in the dataset: 29\% of similes in the train set are neg pairs with explicit negation in the paraphrase but only 12\% of metaphors are, and many similes with as-as structure have implicit negation with \textit{antonym} type pairs which do not exist for metaphors. In Figure \ref{fig:figqa results neg} we show accuracy split by negating and intensifying similes. For the embedding and mid-phrase prompt style (Figure \ref{fig:figqa_results_neg_em}), we see that while older models (GloVe, SBERT) have the expected pattern that performance is significantly weaker on negated sentences ($p<0.05$), Llama models have the opposite pattern.

For the question-answer prompt style, Figure \ref{fig:figqa_results_neg_qa} shows that while accuracy is generally higher, we do see the expected pattern that negated sentences are more difficult for models to interpret. Accuracy for intensifying similes is similarly high across all models but for negating similes is much more varied and mostly lower. This effect is common to different model sizes and architectures: on the train set, all models prompted with the question-answer prompt style have a significantly lower performance on negating similes than intensifying ($p<0.05$) with the exception of Llama-3.1-405B-Instruct. On the test set, we see a very similar pattern of results, although the difference is significant only for GPT-4o-mini (Figure \ref{fig:figqa_results_neg_qtest}, Appendix \ref{appx:test_results}). Note that the test set is around one eighth the size of the train set, which contributes to the lack of statistical significance. Humans showed no significant difference in accuracy between negating and intensifying similes (neg: 0.934, 95\% binomial confidence interval [0.883-0.965], intens: 0.940 [0.894-0.968]).

\begin{figure}[!h]
\centering
    \includegraphics[width=0.8\linewidth]{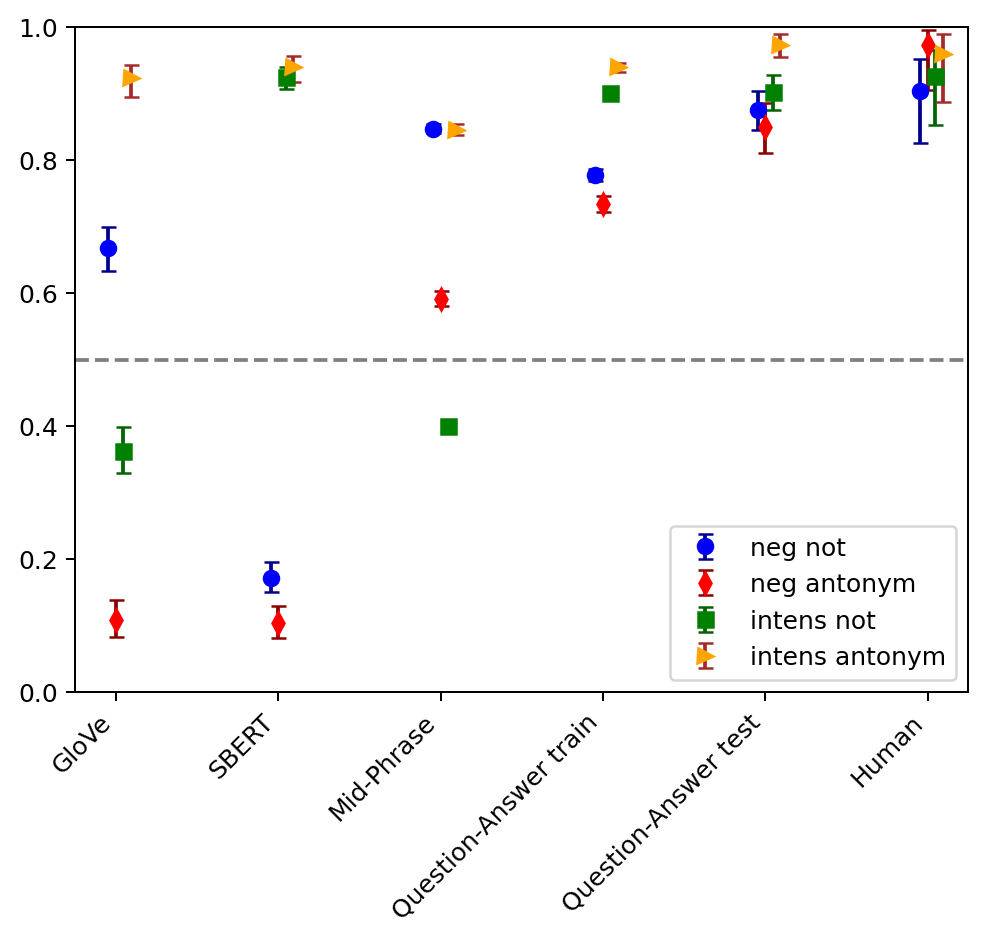}
    \caption{Accuracy on negating and intensifying similes in Fig-QA. Mid-Phrase shows the mean of all Llama3 models; Question-Answer shows the mean of all models using the question-answer method. Bars indicate 95\% bootstrap confidence intervals where an average over models is taken and 95\% binomial confidence intervals otherwise. The dotted line is the random baseline.}
    \label{fig:figqa results neg splits}
\end{figure}
\subsection{Explicit vs Implicit Negation}
In Figure \ref{fig:figqa results neg splits}, we show a breakdown into the \textit{not} and \textit{antonym} categories introduced in Table \ref{tab:simile example} and aggregated across prompting style. 

For the embedding and mid-phrase methods, Figure \ref{fig:figqa results neg splits} shows that the stronger performance of models on negated sentences is not split equally across different negation types. Instead, most models perform more strongly on sentences using explicit negation (\textit{not}) and less strongly on sentences involving antonyms (\textit{antonym}).

For the question-answer prompt style,  Figure \ref{fig:figqa results neg splits} shows that on the train set, performance in \textit{neg} and \textit{intens} sentences is similar across the \textit{not} and \textit{antonym} categories, and fits the expected pattern that \textit{intens} sentences are easier to paraphrase than \textit{neg} sentences. This is in contrast to performance using the mid-phrase and embedding methods, with the exception of SBERT. On the test set, we see a somewhat similar pattern to the train set, although the differences in accuracy between \textit{neg not} and \textit{neg antonym}, and \textit{neg not} and \textit{intens not} are not significant. Humans, however, showed no statistically significant difference in accuracy between the different negation types.

\begin{figure*}[!h]
    \centering
    \begin{subfigure}{0.4\textwidth}
            \includegraphics[width=\textwidth]{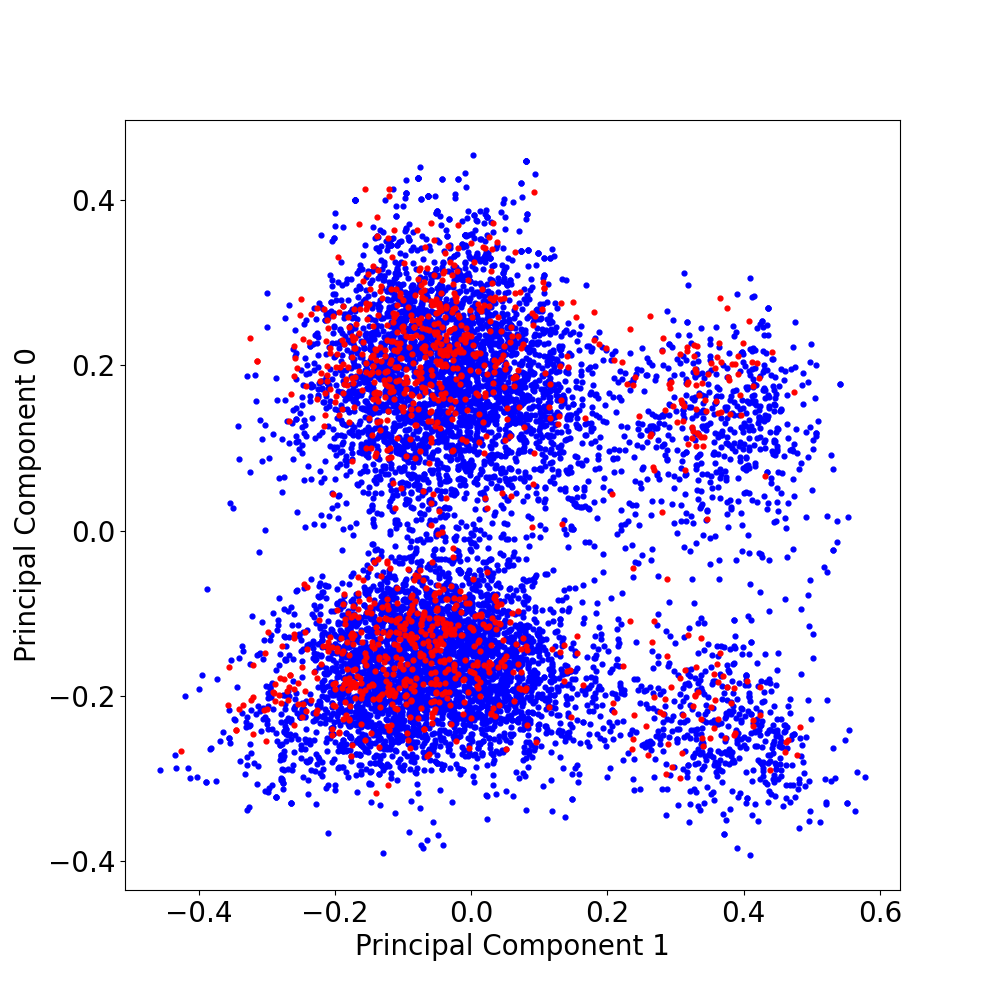}
    \end{subfigure}
    \hspace{1.5cm}
    \begin{subfigure}{0.4\textwidth}
    \includegraphics[width=\textwidth]{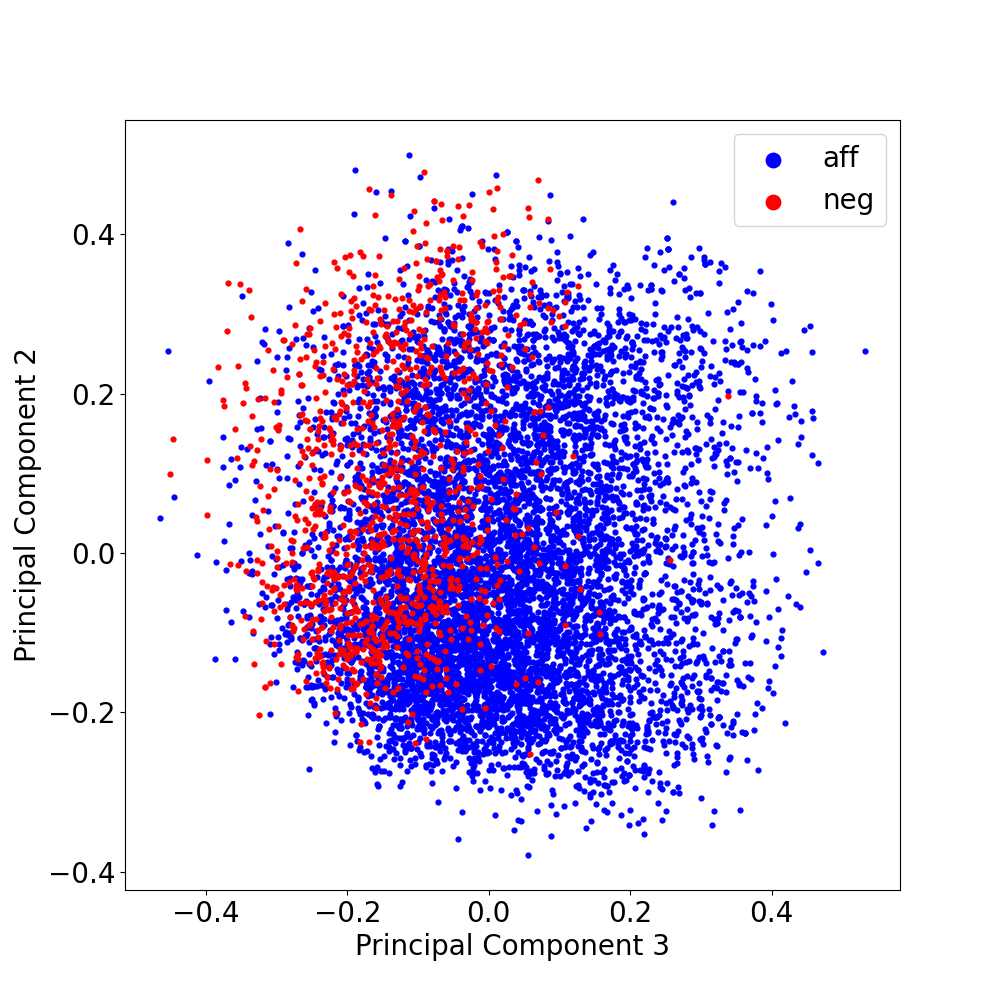}
    \end{subfigure}
    \caption{Plot of principal components 0 to 3 of SBERT sentence embeddings showing negated sentences in red and non-negated sentences in blue.}
    \label{fig:pca SBERT}
\end{figure*}
\section{Embedding Analysis} \label{sbert_analysis}

Multiple factors are likely to contribute to performance on figurative language interpretation. To isolate these we examine the embedding space of SBERT\footnote{SBERT all-MiniLM-L6-v2}. 
We perform Principal Component Analysis (PCA) on the sentence embeddings for the literal paraphrases. Plots of the first four principal components (PCs) are shown in Figure \ref{fig:pca SBERT}. 
PC3 shows a strong separation of negated sentences from non-negated sentences, and there are other clusters which may contribute to performance.

We manually examine sentences from different regions of the first six PCs and observe the following patterns, summarised in Table \ref{tab:pca summary}.

\begin{table}[htbp]
\footnotesize
  \centering
    \begin{tabular}{r|l}
    \multicolumn{1}{l|}{PC} & \multicolumn{1}{c}{SBERT} \\
    \hline
    0     & past/present tense \\
    1     & presence of feminine pronoun or noun  \\
    2     & semantically good/bad \\
    3     & negated/not negated  \\
    4     & concrete/abstract subject \\
    5     & masculine/feminine (both side also contain neuter)  \\
    26    & presence of cats or dogs \\
    \end{tabular}%
    \caption{Summary of the first principal components of SBERT sentence embeddings.}
  \label{tab:pca summary}%
\end{table}%

We assess the effect of each of these categories on paraphrasing performance in the following way. We firstly calculate the value of each sentence embedding on each PC. For each PC, we then take the 10\% of paraphrase embeddings that have the highest value on that PC (labelled `high' in Table \ref{tab:pca_results_acc}), and the 10\% of paraphrase embeddings that have the lowest value on that PC (labelled `low'). We calculate the accuracy over sentences in the high and low categories for each PC, based on the figurative sentences whose apt paraphrases are in the high and low categories. We go on to look at the correlation of each PC with negation. Table \ref{tab:pca_results_acc} gives the accuracy and correlation with negation for each PC.

\begin{table}[htbp]
    \centering
    \footnotesize
    \begin{tabular}{r|rr|rr}
    \multicolumn{1}{l|}{PC} & \multicolumn{2}{c|}{accuracy (all)} & \multicolumn{2}{c}{prop. neg.} \\
    \multicolumn{1}{l|}{all} & \cellcolor[rgb]{ .988,  .988,  1}0.5735 &       & \cellcolor[rgb]{ .988,  .988,  1}0.1220 &  \\
          & \multicolumn{1}{l}{low} & \multicolumn{1}{l|}{high} & \multicolumn{1}{l}{low} & \multicolumn{1}{l}{high}\\
    0     & \cellcolor[rgb]{ .796,  .851,  .933}0.7047 & \cellcolor[rgb]{ .984,  .898,  .91}0.4852 & \cellcolor[rgb]{ .71,  .831,  .635}0.0604 & \cellcolor[rgb]{ .988,  .965,  .961}0.1647 \\
    1     & \cellcolor[rgb]{ .984,  .945,  .957}0.5335 & \cellcolor[rgb]{ .937,  .953,  .984}0.6081 & \cellcolor[rgb]{ .988,  .969,  .965}0.1625 & \cellcolor[rgb]{ .941,  .961,  .937}0.1120 \\
    2     & \cellcolor[rgb]{ .729,  .804,  .91}0.7497 & \cellcolor[rgb]{ .984,  .843,  .855}0.4325 & \cellcolor[rgb]{ .486,  .706,  .341}0.0110 & \cellcolor[rgb]{ .98,  .914,  .875}0.2602  \\
    3     & \cellcolor[rgb]{ .976,  .635,  .643}0.2239 & \cellcolor[rgb]{ .616,  .725,  .871}0.8255 &  \cellcolor[rgb]{ .973,  .827,  .737}0.4083 & \cellcolor[rgb]{ .439,  .678,  .278}0.0000  \\
    4     & \cellcolor[rgb]{ .851,  .89,  .953}0.6674 & \cellcolor[rgb]{ .984,  .863,  .871}0.4490 &  \cellcolor[rgb]{ .529,  .729,  .4}0.0209 & \cellcolor[rgb]{ .984,  .925,  .894}0.2382  \\
    5     & \cellcolor[rgb]{ .984,  .922,  .933}0.5093 & \cellcolor[rgb]{ .922,  .941,  .976}0.6202 &  \cellcolor[rgb]{ .988,  .969,  .969}0.1581 & \cellcolor[rgb]{ .635,  .788,  .537}0.0439  \\
    \end{tabular}%
  \caption{Leftmost columns: accuracy for sentences that take the 10\% lowest and highest values on the first 6 PCs. Red indicates accuracy lower than 0.5 baseline and blue indicates higher accuracy. Rightmost columns: proportion of negated sentences in each split. Green indicates a lower proportion of sentences with negation than the overall dataset and orange indicates higher.}
  \label{tab:pca_results_acc}%
\end{table}%

\begin{figure*}
    \centering
    \begin{subfigure}{0.4\textwidth}
        \includegraphics[height=0.25\textheight]{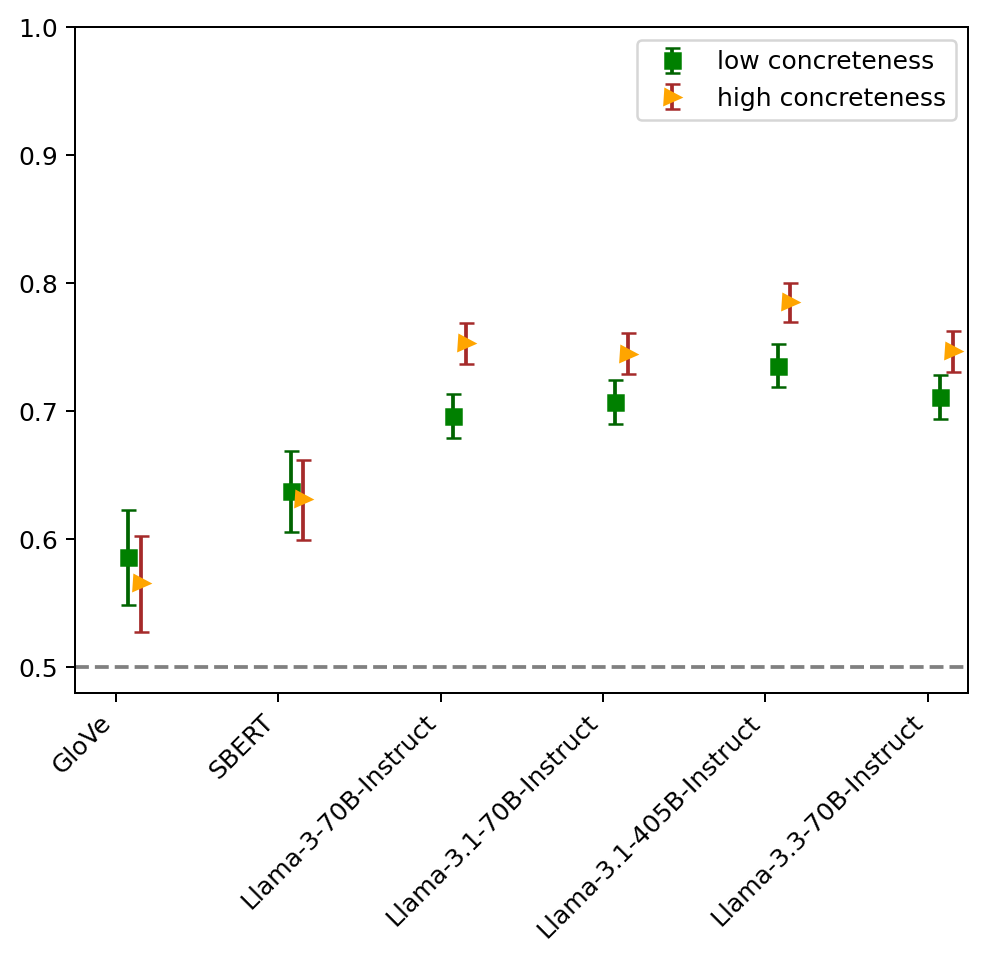}
        \caption{Embedding and Mid-Phrase}
    \end{subfigure}
    \hspace{1cm}
    \begin{subfigure}{0.4\textwidth}
     \includegraphics[height=0.25\textheight]{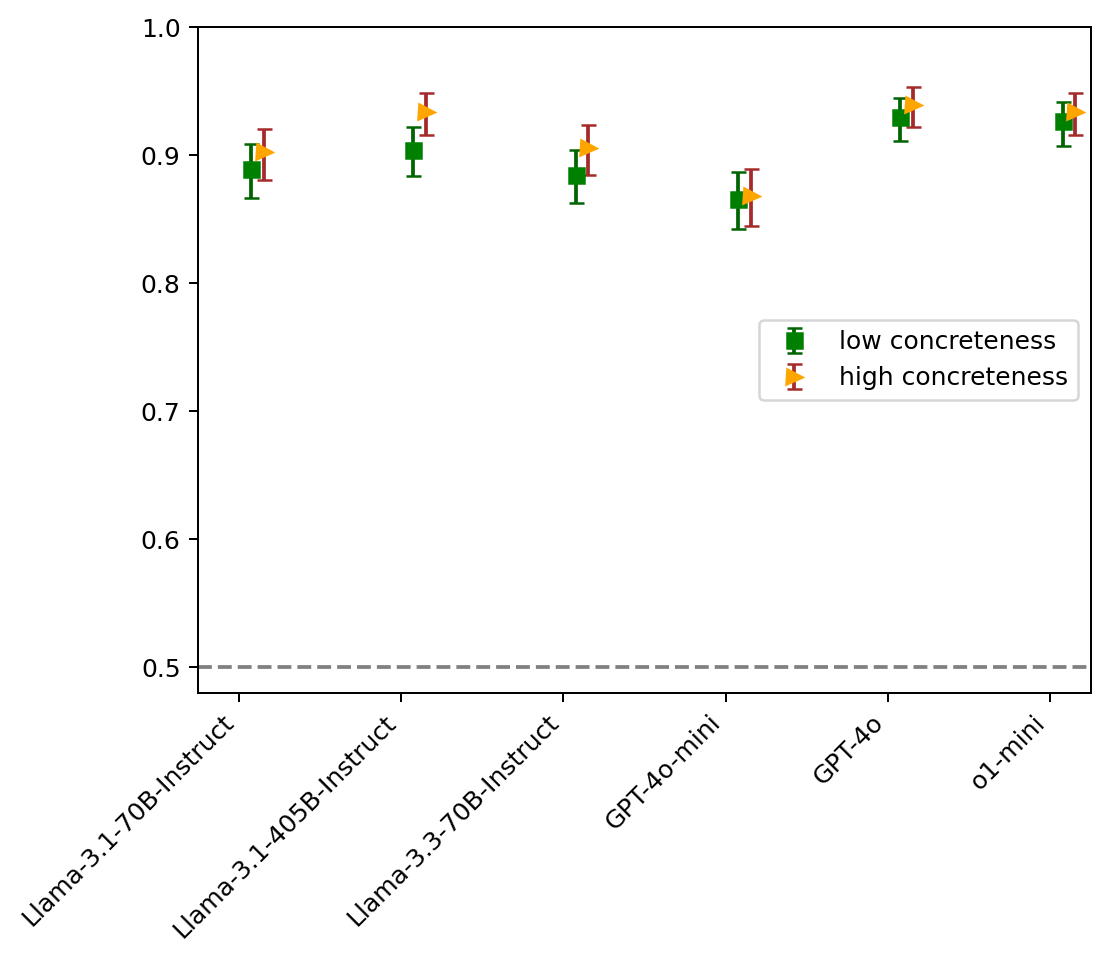}
     \caption{Question-Answer}
    \end{subfigure}
    \caption{Model accuracy for 10\% of sentences with highest and lowest concreteness rating in the Fig-QA train set using (a) the embedding and mid-phrase methods on the train set and (b) the question-answer method. The bars indicate 95\% binomial confidence intervals except Llama models in (a) which are 95\% bootstrap confidence intervals. The dotted line is the random baseline.}
    \label{fig:concreteness acc}
\end{figure*}

PC3 (corresponding to negation) shows the strongest relationship with model accuracy, supporting the previous hypothesis that models struggle with negation in particular. PC0 (tense), PC2 (semantically good/bad) and PC4 (concreteness) were also shown to have a strong effect on accuracy. In addition to PC3, many of the components show some level of correlation with negation, showing that each PC does not correlate to just one aspect of language. These results point to various factors having an effect on accuracy, and that these factors are intertwined with negation. We assess the effect of concreteness and tense on model accuracy, reported in the following section. 

\section{Concreteness and Tense}

\begin{figure*}[t]
    \centering
    \begin{subfigure}{0.4\textwidth}
    \includegraphics[height=0.25\textheight]{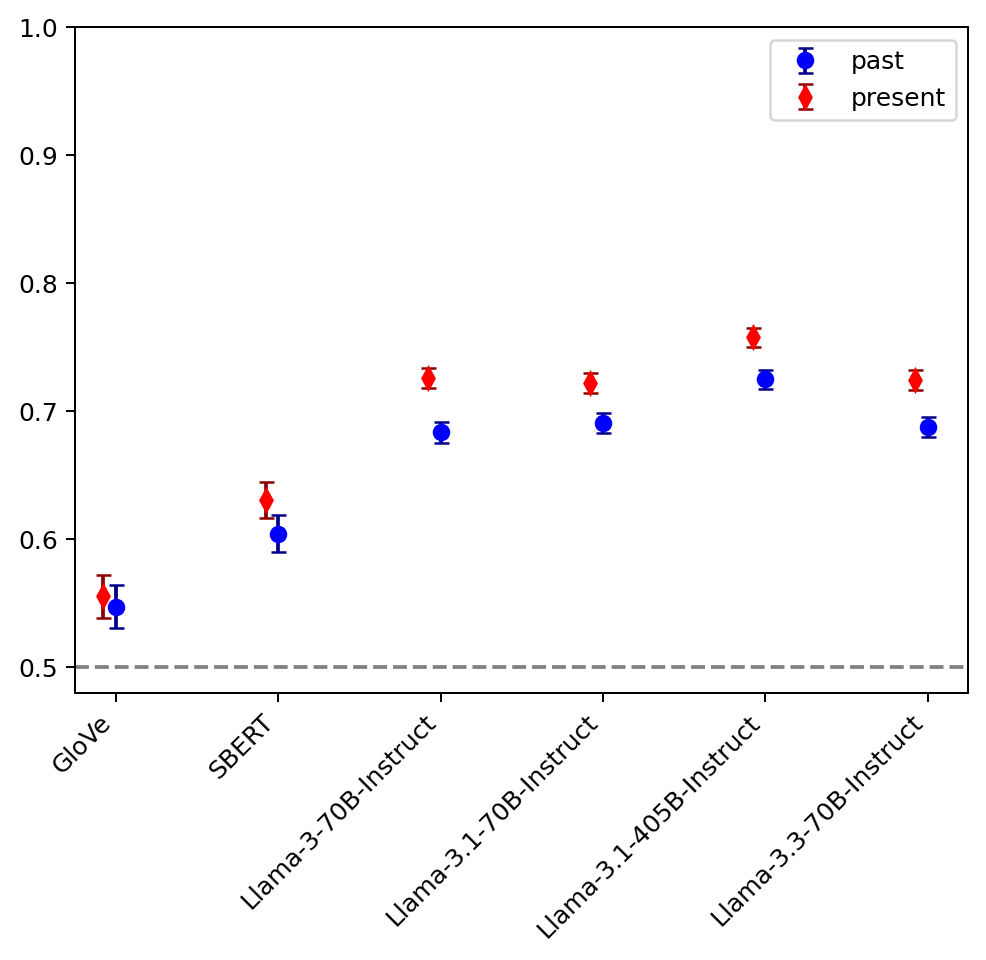} 
    \caption{Embedding and Mid-Phrase}
    \end{subfigure}
    \hspace{1cm}
    \begin{subfigure}{0.4\textwidth}
    \includegraphics[height=0.25\textheight]{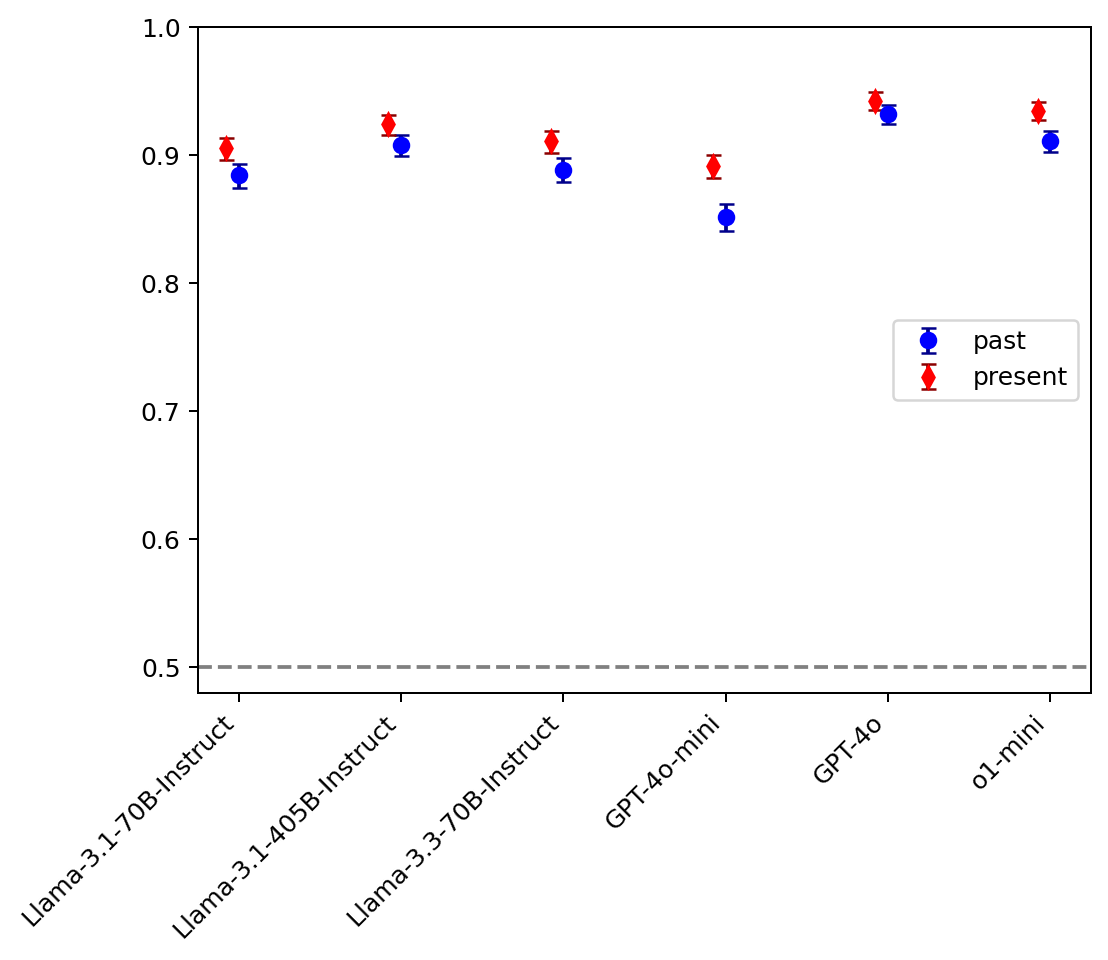}
    \caption{Question-Answer}
    \end{subfigure}
    \caption{Model accuracy for sentences in the Fig-QA train set in past and present tense using (a) the embedding and mid-phrase methods on the train set and (b) the question-answer method. The bars indicate 95\% binomial confidence intervals except Llama models in (a) which are 95\% bootstrap confidence intervals. The dotted line is the random baseline.}
    \label{fig:tense acc}
\end{figure*} 

\subsection{Dataset}
We annotate Fig-QA with labels for concreteness and tense. We use SpaCy \citep{spacy} to extract tense labels (Past or Pres) for the paraphrases. Of the 9,110 instances of the train set, there are 160 instances without a tense label; these are mostly where the paraphrase has the auxiliary verb ``can'', ``could'', ``will'' or ``would''. We labelled 4,494 sentences as \textit{Past} and 4,456 as \textit{Pres}.

To assign concreteness ratings, we again use spaCy to extract the subject of each paraphrase. We then match the subject to a pre-compiled list of words with concreteness annotations \citep{brysbaert_13}, where more concrete words are those that refer to perceptually salient objects, e.g.\ \textit{chair} vs \textit{derivative}. 203 instances lack concreteness ratings, due either to the paraphrase lacking a subject, or to the subject not being found in the concreteness list - mostly where the subject was a proper name. The sentences were then grouped into highest and lowest 10\%, giving 911 sentences for each of high and low concreteness.

\subsection{Concreteness}
We compute accuracy across the high and low concreteness splits, presented in Figure \ref{fig:concreteness acc}.

We see that concreteness has only a small effect on interpretation accuracy, particularly on the question-answer prompt style, where none of the differences were statistically significant on the train set or the test set (Figure \ref{fig:figqa_results_concreteness_qtest}, Appendix \ref{appx:test_results}. For the mid-phrase prompt style on the train set, we see that Llama models perform slightly better on more concrete sentences with this difference being significant for all Llama models tested. Humans had a lower performance on more concrete sentences, but this was not statistically significant ($p=0.38$) (low concreteness: 0.956, 95\% binomial confidence interval [0.900-0.983], high concreteness: 0.947 [0.887-0.977]).
The small effect of concreteness, paired with the correlation of PC4 with negation, suggests that the difference in accuracy across PC4 was due to negation and not concreteness.

\subsection{Tense}
To check the effect of tense, we split the dataset by tense annotation (past or present), and calculate accuracy on each split. Results are shown in Figure \ref{fig:tense acc}.

We see that tense has a small effect on model accuracy, with all models performing slightly better at present tense sentences than past tense on the train set. This is the case with both the mid-phrase method and the question-answer method. This difference was significant for all Llama models using the mid-phrase method and all models with the question-answer method. On the test set, no models showed a significant difference in accuracy between past and present tense, including humans (past: 0.952, 95\% binomial confidence interval [0.929-0.968], present: 0.941 [0.919-0.957]), Figure \ref{fig:figqa_results_concreteness_qtest}, Appendix \ref{appx:test_results}.
Again, the difference in accuracy of SBERT across PC0 appears to be mainly due to negation.

\begin{table*}
  \centering
    \footnotesize
    \begin{tabular}{ll|ll}
    Original Sentence & Original Interpretation & Literal Sentence & Literal Interpretation \\
    \hline
    This is as fun as a bowl of nachos & It’s very fun & This is entertaining & It’s fun \\
    This is as fun as stubbing your toe & It’s not fun at all & This is boring & It’s not fun \\
    That man has the honesty of a saint & The man is very honest & That man is truthful & The man is honest \\
    That man has the honesty of a crook & The man is not honest at all & That man is deceitful & The man is not honest \\
    \end{tabular}%
    \caption{Example of two sentence pairs with negation from Fig-QA, and in the new literal dataset}
  \label{tab:litneg example}%
\end{table*}%
\section{Interaction of Negation and Figurativeness}
We have seen that the presence of negation affects the performance of models in interpreting figurative language. To determine whether the difference in performance is due only to negation, or whether there is an interaction between negation and figurative language, we develop a small literal negation dataset based on the Fig-QA dataset.

\begin{figure*}
    \centering
    \begin{subfigure}{0.4\textwidth}
        \includegraphics[height=0.25\textheight]{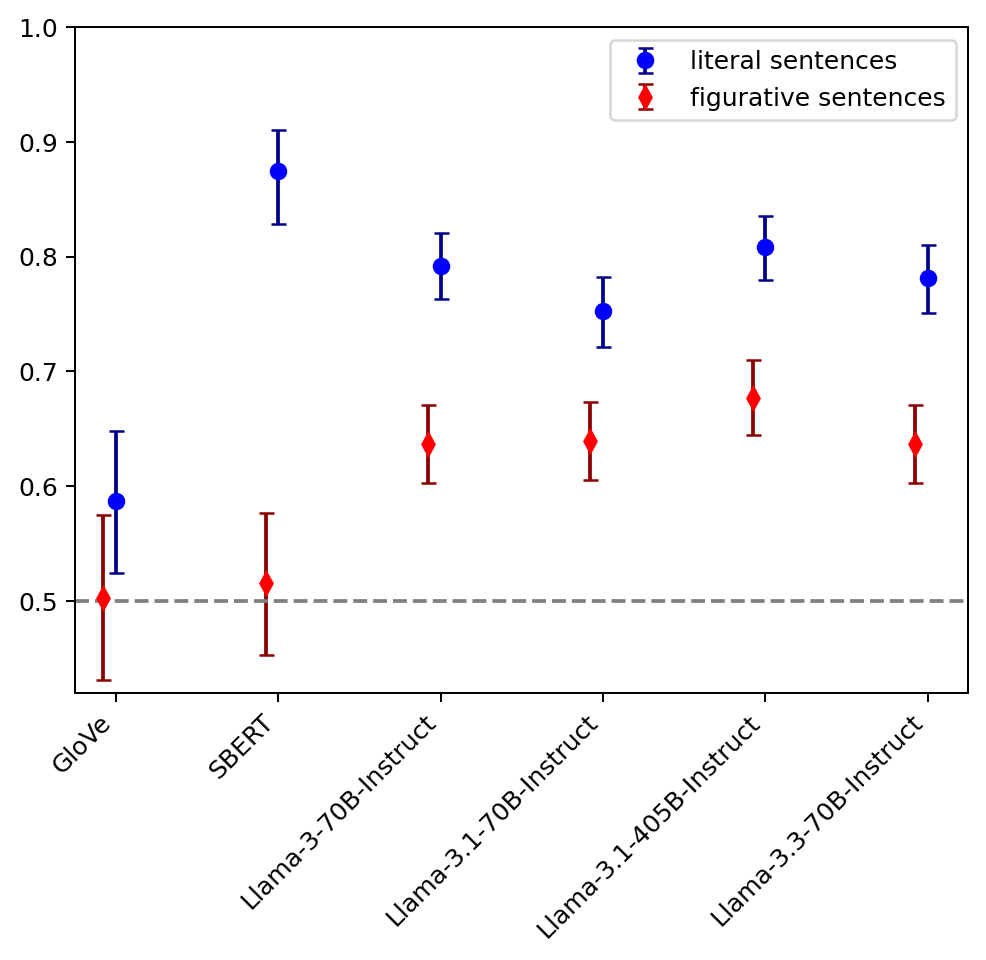} 
        \caption{Embedding and Mid-Phrase}
    \end{subfigure}
        \hspace{1cm}
    \begin{subfigure}{0.4\textwidth}
        \includegraphics[height=0.25\textheight]{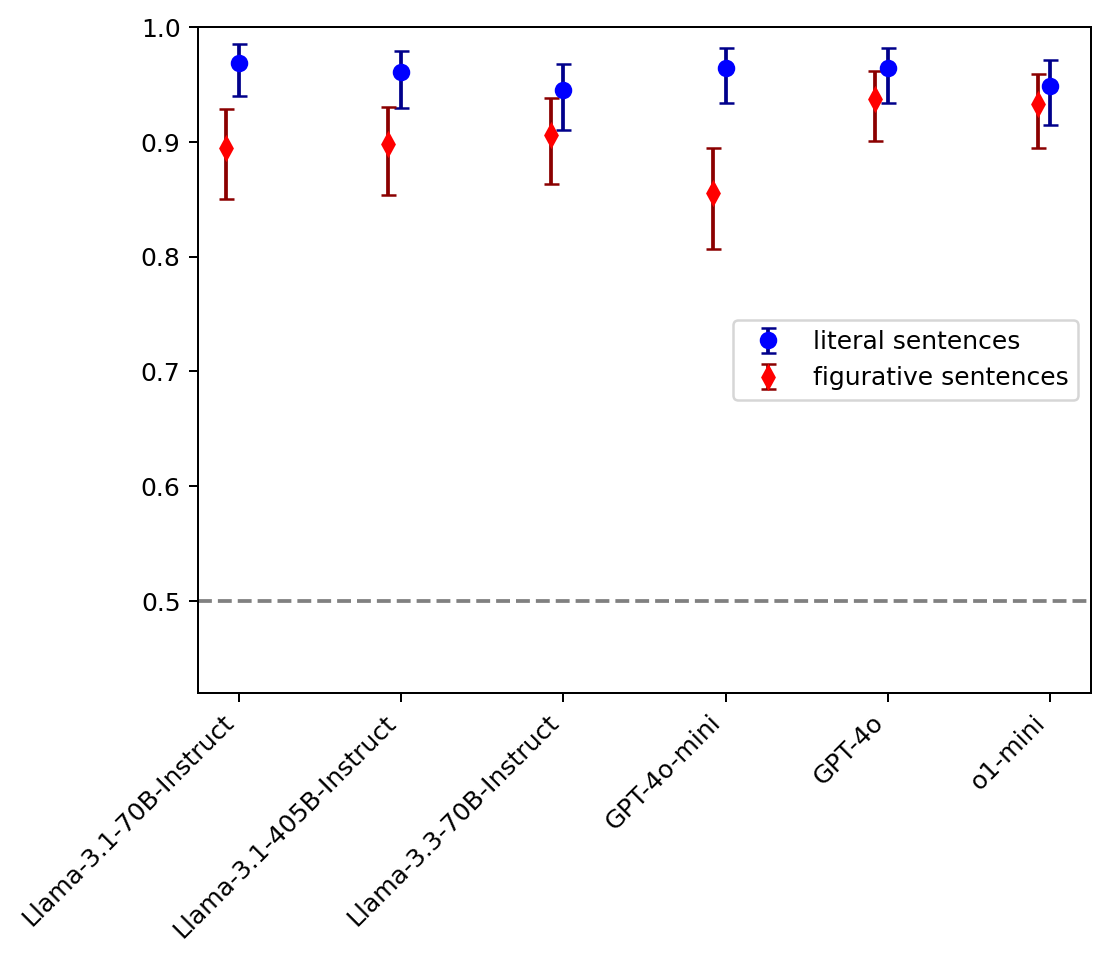}
        \caption{Question-Answer}
    \end{subfigure}
    \caption{Model accuracy for the literal negation dataset and the corresponding figurative sentences from Fig-QA using (a) the embedding and mid-phrase methods and (b) the question-answer method. The bars indicate 95\% binomial confidence intervals except Llama models in (a) which are 95\% bootstrap confidence intervals. The dotted line is the random baseline.}
    \label{fig:litneg results}
\end{figure*}

\subsection{Literal Negation Dataset}
We took a random sample of 128 figurative sentence pairs from Fig-QA, selecting pairs containing one negating and one intensifying simile. Using WordNet and online dictionary resources, we manually wrote literal versions of the figurative sentences, using a synonym and antonym of the adjective in the simile, resulting in 256 sentences with paraphrases. The paraphrases were simplified such that the only difference between the two options is the negation word. For example, for the figurative sentence, \textit{He's as busy as a sloth}, the original paraphrase was \textit{He's lazy, not busy at all}. This was simplified to \textit{He's not busy}.

The dataset was written by one of the authors and manually validated by the other two and two independent observers, ensuring that sentences followed the correct format and were labelled correctly.
Table \ref{tab:litneg example} shows an example of an original sentence pair from Fig-QA and the corresponding literal pair.

\subsection{Results}
Figure \ref{fig:litneg results} shows accuracy on the new literal sentences as well as the 256 corresponding figurative sentences from Fig-QA. Across all models and prompt styles, performance on literal sentences is greater than or equal to performance on figurative sentences.

Figure \ref{fig:litneg results neg} shows the results split between negating and intensifying sentences, aggregated in the same way as Figure \ref{fig:figqa results neg splits}. With the mid-phrase method, models had higher accuracy on \textit{neg} than \textit{intens} sentences, as also observed 
in Figure \ref{fig:figqa results neg}. We also see that within the \textit{neg} and \textit{intens} categories, performance on literal sentences is stronger.

Using the question-answer method, models generally have high performance. The models show similarly high accuracy for the different sentence groups except for figurative \textit{neg} sentences which is lower.
Models tested using the question-answer method are able to interpret negation in literal sentences, but they find the interaction between figurativeness and negation particularly challenging to interpret.

\section{Discussion and Conclusions}
\begin{figure}[h!]
    \centering
    \includegraphics[width=0.8\linewidth]{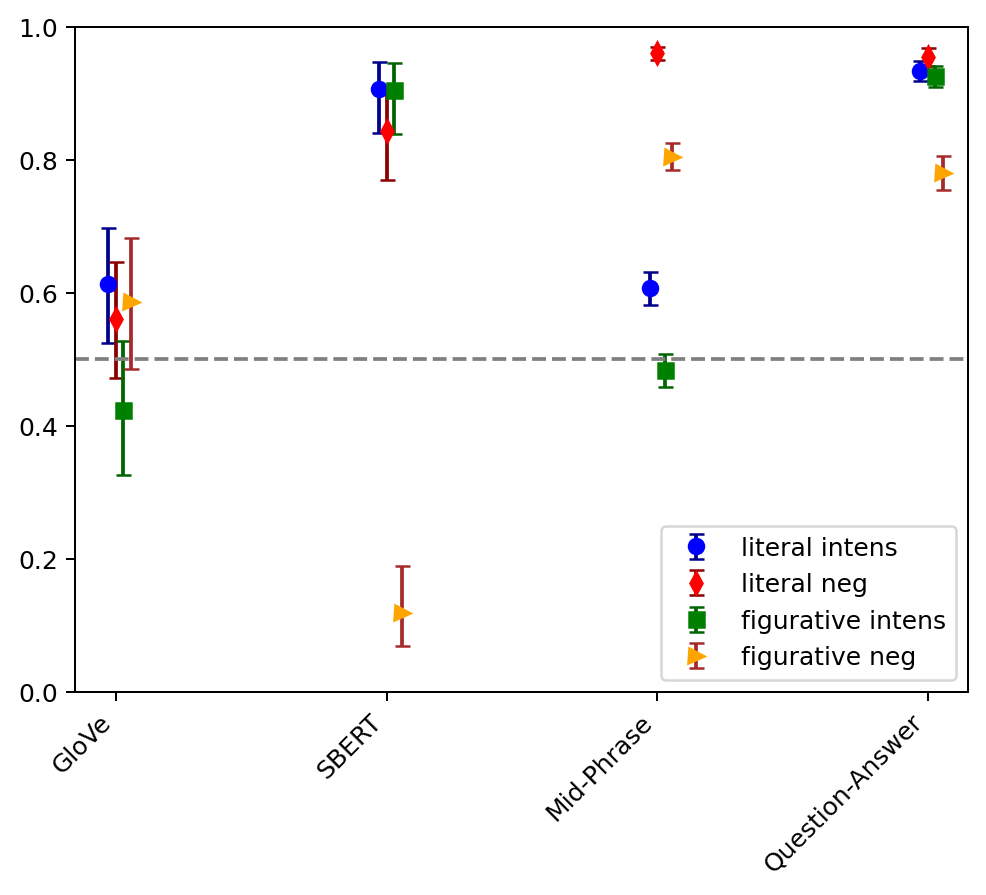}
    \caption{Model accuracy for the literal negation dataset and the corresponding figurative sentences from Fig-QA split into negating and intensifying sentences. 
    Bars indicate 95\% bootstrap confidence intervals where an average over models is taken and 95\% binomial confidence intervals otherwise. The dotted line is the random baseline.}
    \label{fig:litneg results neg}
\end{figure}
Figurative language and negation have each been shown to be challenging for LLMs. In this work, we investigated performance on their combination, using the Fig-QA dataset. 

Our overall findings are that the combination of figurative language and negation, in particular simile, does indeed present a challenge to LLMs. In particular, there is significant difference in performance across most models in processing similes involving negation (Figure \ref{fig:figqa results neg}). In considering different types of negation (explicit use of the word `not' vs use of an antonym), patterns of model performance show differences across sentence types. (Figure \ref{fig:figqa results neg splits}). Differences in performance on negated vs intensifying similes, and on subtypes of negated similes are not seen in humans on the test split of the data. Models' performance on the test split does indeed show these differences, although the differences are not significant, due to the smaller size of the dataset (334 neg pairs on the test set vs 3302 on the train set).

These differences in performance are not exhibited across concreteness, although we do see a small difference in tense. Furthermore, for the most up-to-date models, the combination of figurative language and negation results in particularly low performance, when contrasted with either negation or figurative language on their own (Figure \ref{fig:litneg results neg}, Question-Answer method).

However, another key finding is that the method of eliciting responses from models has a large effect. While the question-answer method generally yielded more accurate results, it could be considered a further task from the real use case than the mid-phrase method, since in most cases the model would not get to see the two options. When we directly assess the phrase likelihood of different paraphrases (Figure \ref{fig:figqa_results_neg_em}), results are strikingly different to performance when using the question-answer prompting method (Figure \ref{fig:figqa_results_neg_qa}). We conjecture that models may have seen negated similes in the form of irony during training, and that this may have improved performance.

Overall, we find that the combination of negation and figurative language presents a challenge even to modern LLMs, and that prompting methods have a large effect on performance. We might say that LLMs should be fine-tuned to understand the combination of negation and figurative language, however this is a piecemeal solution. A truly flexible and creative language user---like a human---can understand these novel combinations of phenomena without special training.

It is also important to consider how language models may be used in downstream applications. We have seen that prompting styles make a big difference to performance, but that the prompting style with highest performance is one which is arguably less true to how linguistic phenomena may occur in the real world. We therefore argue that thought should be given to how LLM performance should be assessed with respect to their use in real-world scenarios.

\paragraph{Limitations and Future Work}
The fact that using different kinds of prompts for the same task can give drastically different results can give a deeper insight into how models interpret figurative language which could prompt further research.

The new literal dataset only contains 256 sentences; a larger dataset would allow for further analysis breaking down negation types.

Human annotations were not available for the entire dataset. A fully annotated dataset could allow further in-depth analysis into which sentences are difficult for models and humans to interpret.

\section*{Acknowledgements}
We are grateful to our action editor Hinrich Schütze, and the anonymous reviewers for their comprehensive and constructive feedback which strengthened our work.
Thanks to Beth Pearson (University of Bristol) and Luan Fletcher (ILLC, University of Amsterdam) for providing annotations on the literal dataset.
This work was carried out using the computational facilities of the Advanced Computing Research Centre, University of Bristol - http://www.bristol.ac.uk/acrc/.
This research was supported by funding from the Engineering and Physical Sciences Research Council.

\bibliography{tacl2021}
\bibliographystyle{acl_natbib}

\clearpage

\appendix

\section{Results for Individual Mid-Phrases} \label{appx:midphrase}

\begin{figure}[h!]
    \centering
    \includegraphics[width=0.95\linewidth]{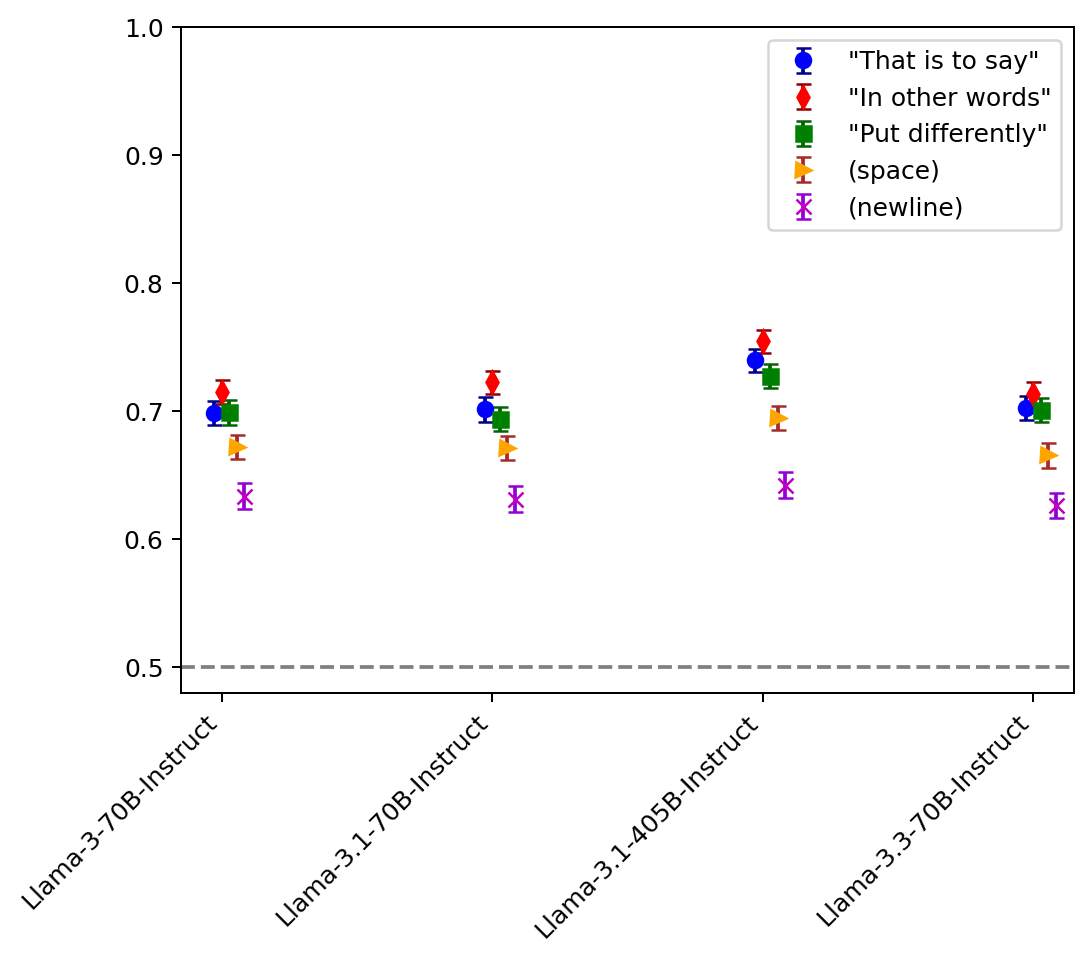}
    \caption{Model accuracy for the Fig-QA train set using different mid-phrases. Bars indicate 95\% binomial confidence intervals. The dotted line is the random baseline.}
    \label{fig:midphrase results}
\end{figure}

\section{Test Set Results}
\label{appx:test_results}

\begin{figure}[H]
\includegraphics[width=0.95\linewidth]{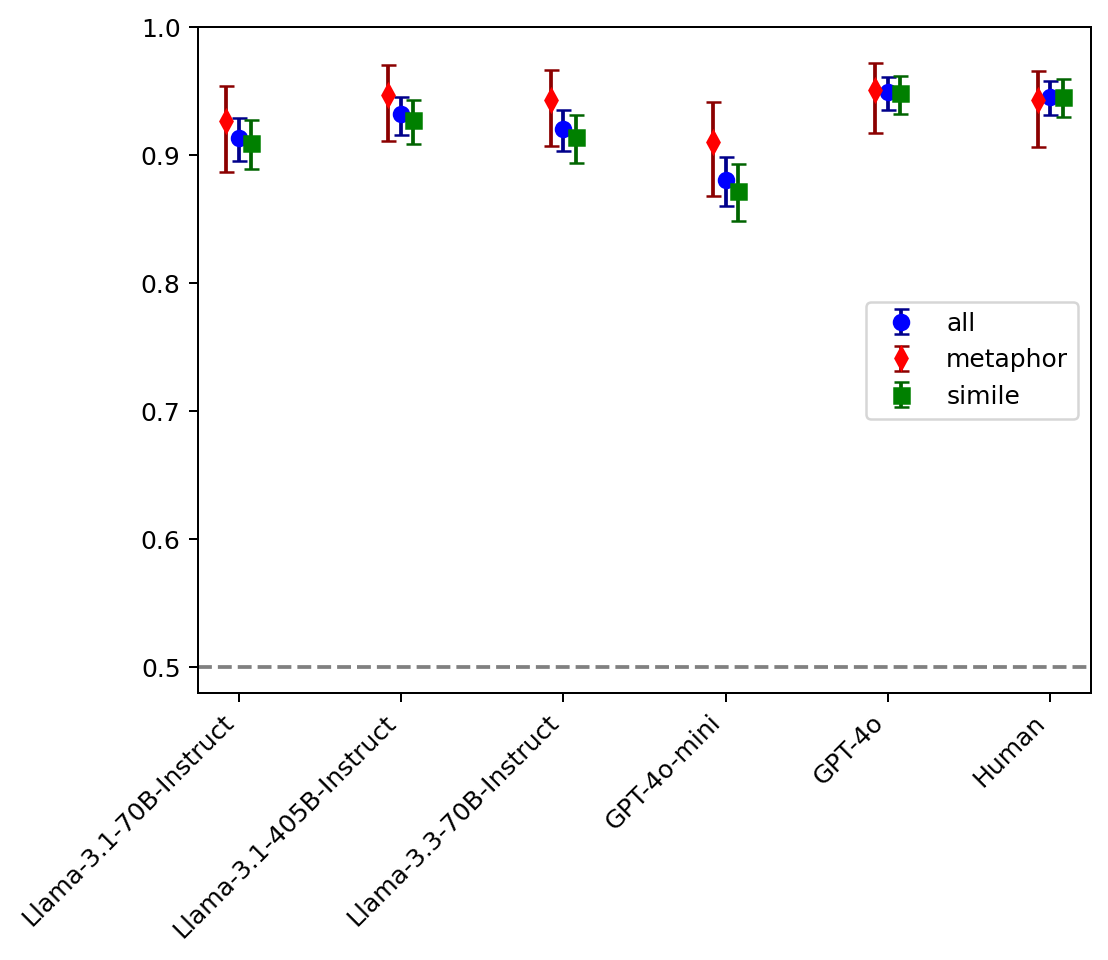}
    \caption{Accuracy of models on the Fig-QA test set using the question-answer method. The bars indicate 95\% binomial confidence intervals. The dotted line is the random baseline.}
    \label{fig:figqa_results_overall_qtest}
\end{figure}

\begin{figure}[!h]
\includegraphics[width=0.95\linewidth]{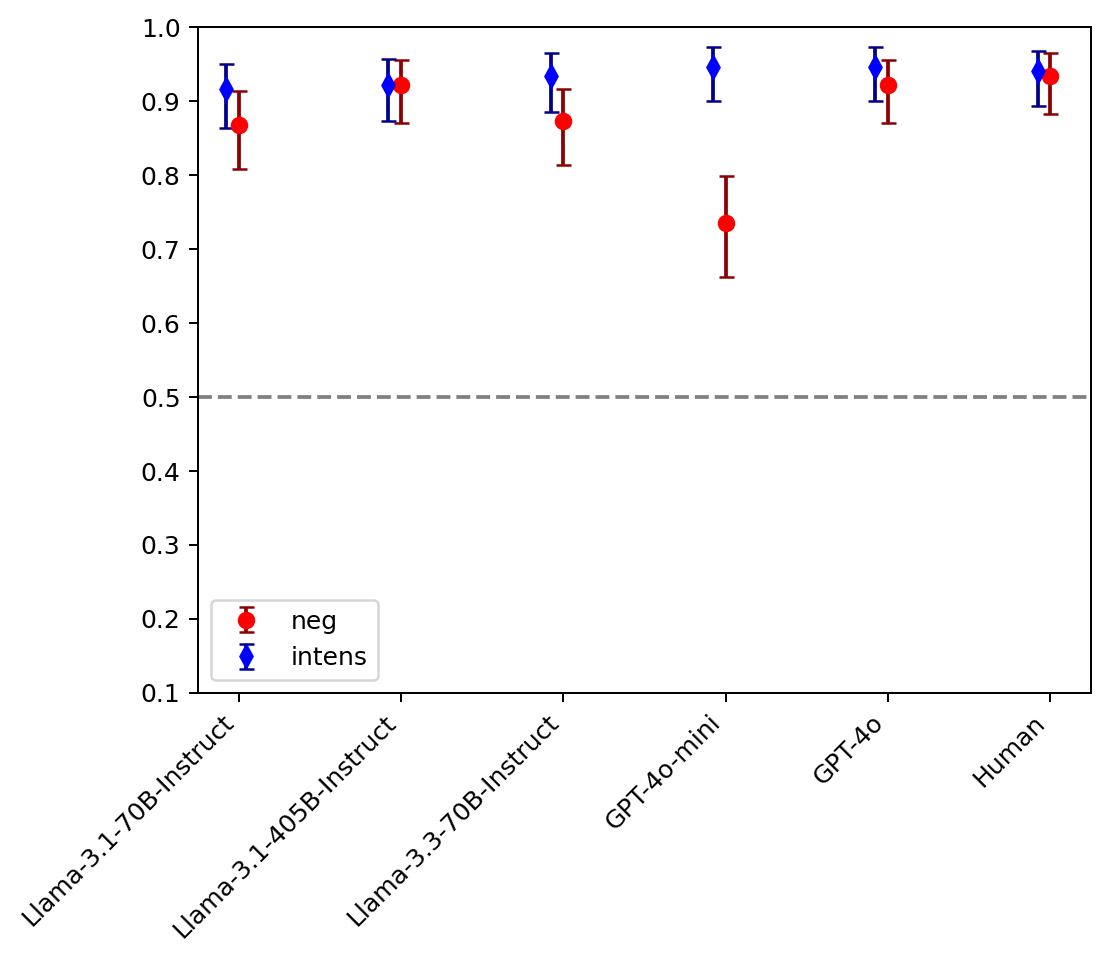}
    \caption{Accuracy of models on negating and intensifying similes in the Fig-QA test set using the question-answer method. The bars indicate 95\% binomial confidence intervals. The dotted line is the random baseline.}
    \label{fig:figqa_results_neg_qtest}
\end{figure}

\begin{figure}[!h]
\includegraphics[width=0.95\linewidth]{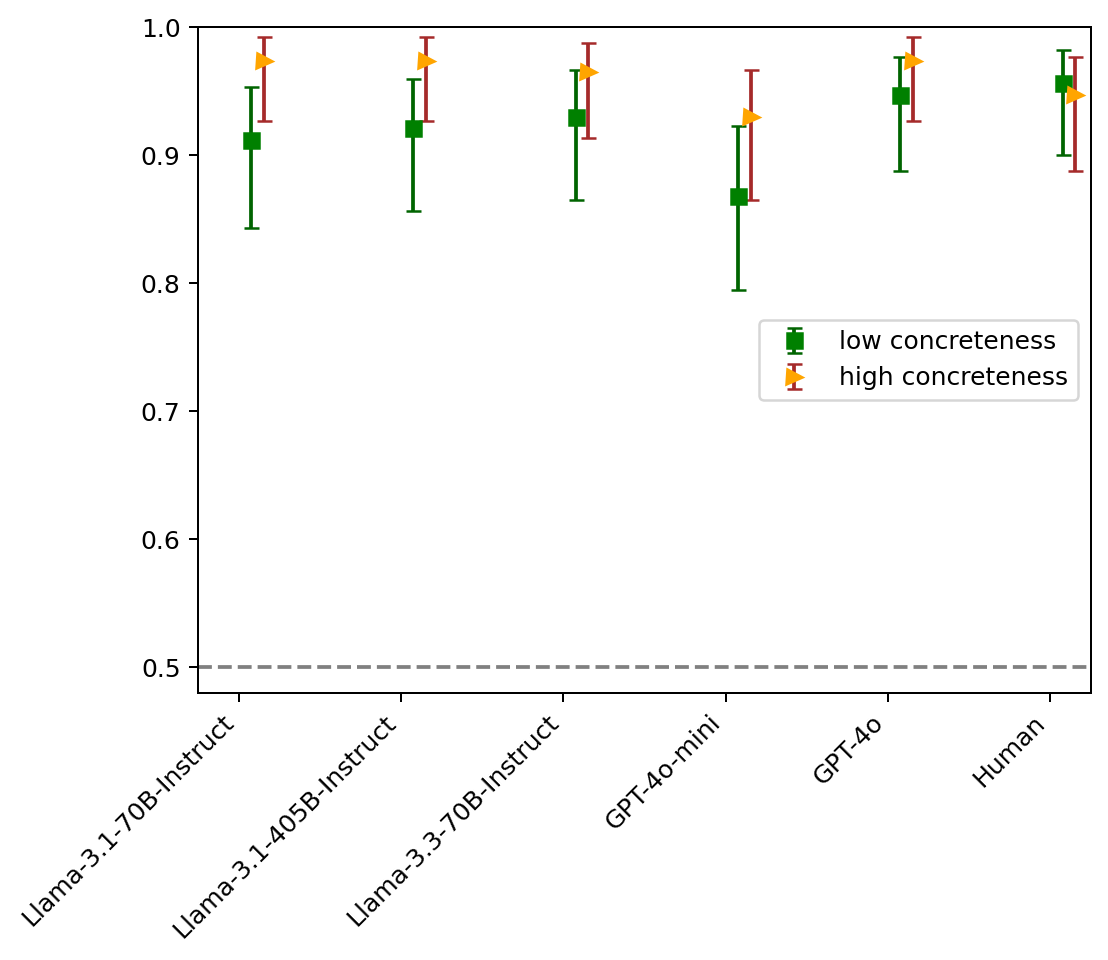}
    \caption{Accuracy of models on sentences with high and low concreteness in the Fig-QA test set using the question-answer method. The bars indicate 95\% binomial confidence intervals. The dotted line is the random baseline.}
    \label{fig:figqa_results_concreteness_qtest}
\end{figure}

\begin{figure}[!h]
\includegraphics[width=0.95\linewidth]{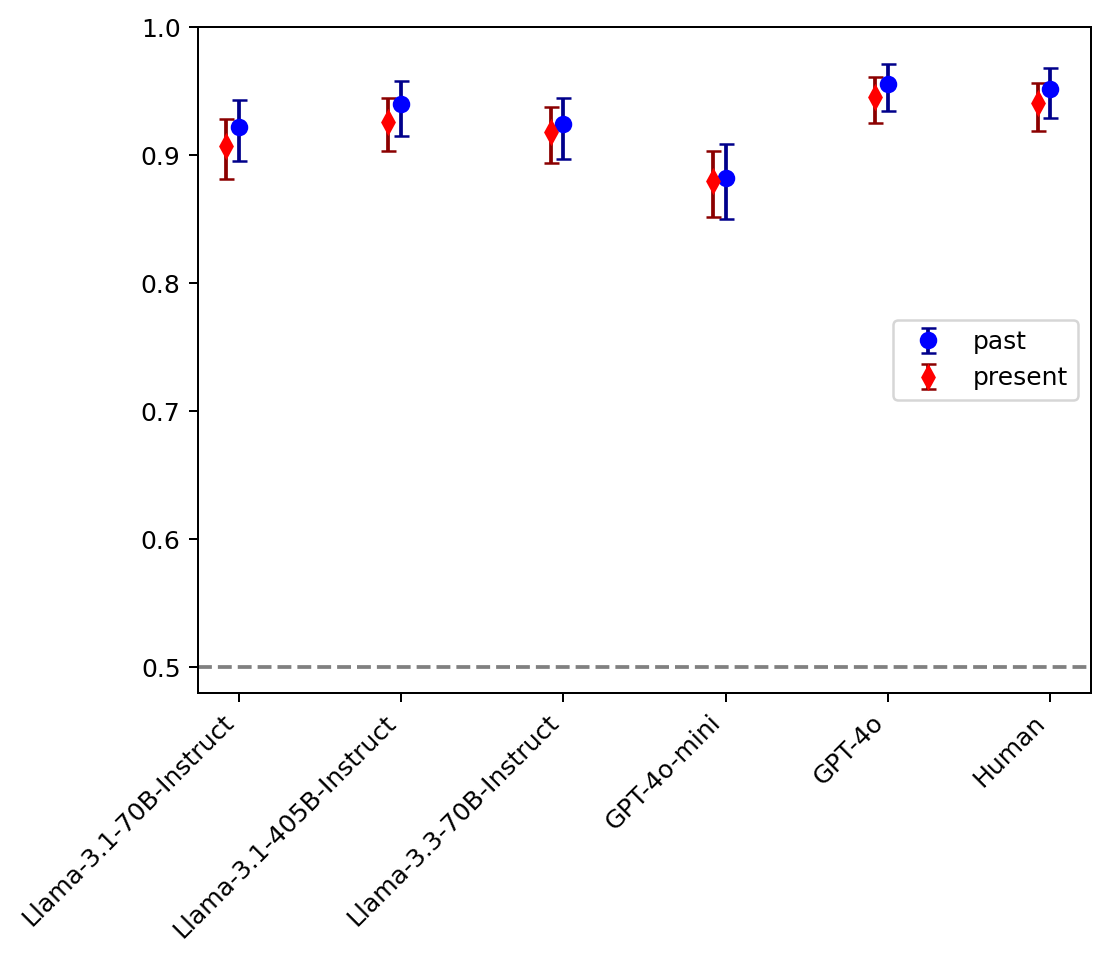}
    \caption{Accuracy of models on sentences with past and present tense in the Fig-QA test set using the question-answer method. The bars indicate 95\% binomial confidence intervals. The dotted line is the random baseline.}
    \label{fig:figqa_results_tense_qtest}
\end{figure}

\section{Original Mid-Phrase Results}
\label{appx:original_results}

\begin{figure}[H]
\includegraphics[width=0.9\linewidth]{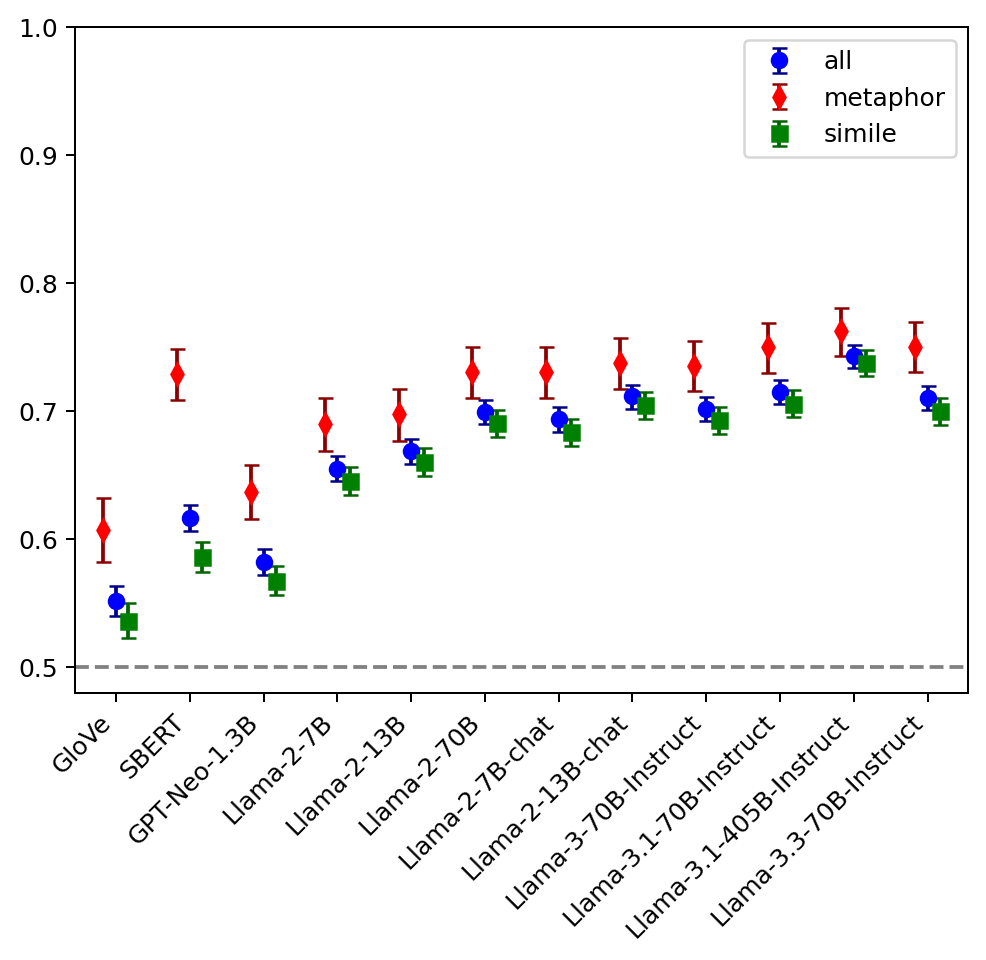}
    \caption{Accuracy of models on the Fig-QA train set using the embedding and mid-phrase method with ``That is to say,''. The bars indicate 95\% binomial confidence intervals. The dotted line is the random baseline.}
    \label{fig:figqa_results_overall_em original}
\end{figure}

\begin{figure}[!h]
\includegraphics[width=0.9\linewidth]{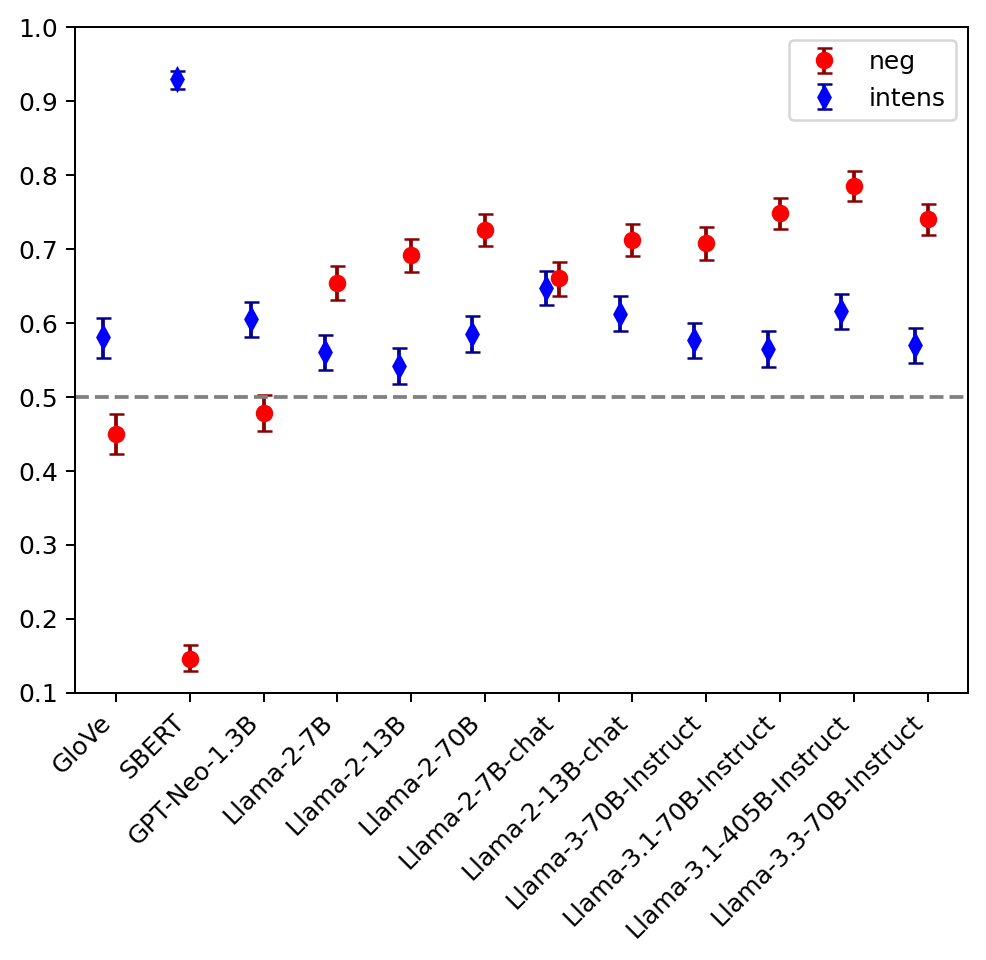}
    \caption{Accuracy of models on negating and intensifying similes in the Fig-QA train set using the embedding and mid-phrase method with ``That is to say,''. The bars indicate 95\% binomial confidence intervals. The dotted line is the random baseline.}
    \label{fig:figqa_results_neg_em original}
\end{figure}

\begin{figure}[!h]
\includegraphics[width=0.9\linewidth]{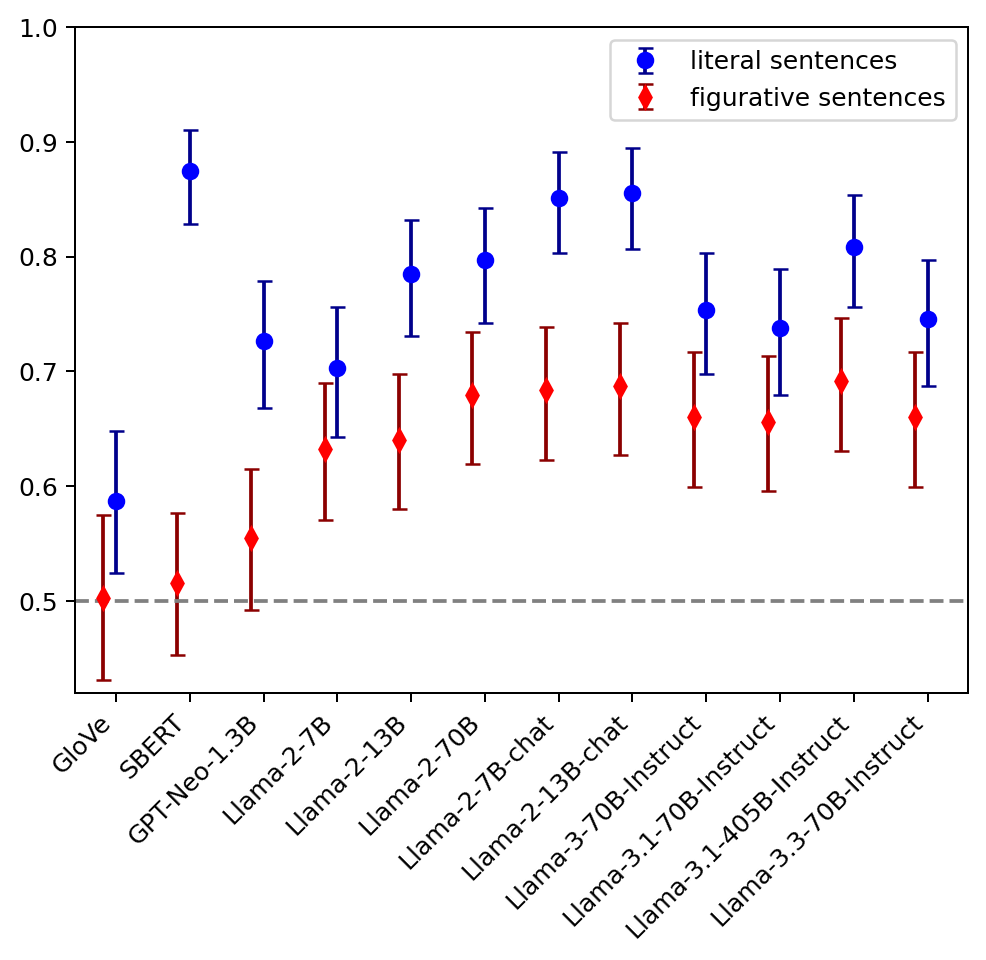} 
        \caption{Model accuracy for the literal negation dataset and the corresponding figurative sentences from Fig-QA using the embedding and mid-phrase method with ``That is to say,''. The bars indicate 95\% binomial confidence intervals. The dotted line is the random baseline.}
    \label{fig:litneg results original}
\end{figure}

\end{document}